\documentclass[journal]{IEEEtran}

\ifCLASSINFOpdf
\else
\fi

\usepackage{graphicx}
\usepackage{amssymb}
\usepackage{epstopdf}
\usepackage{subfigure}
\usepackage{algorithm}
\usepackage{algorithmic}
\usepackage{amsmath}
\usepackage{enumerate}
\usepackage{color}
\usepackage{textcomp}
\usepackage{amsthm}
\usepackage{relsize}
\usepackage{cite}

\newcommand{\M}{\ensuremath{\mathcal{M}}}
\newcommand{\Hs}{\ensuremath{\mathcal{H}}}
\newcommand{\E}{\ensuremath{\mathcal{E}}}
\newcommand{\Ev}{\ensuremath{\mathbb{E}}}
\newcommand{\R}{\ensuremath{\mathbb{R}}}

\newcommand{\Rn}{\ensuremath{\mathbb{R}^n}}
\newcommand{\Rd}{\ensuremath{\mathbb{R}^d}}

\newcommand{\tr}{\ensuremath{ \text{tr}}}
\newcommand{\poly}{\ensuremath{ \text{poly}}}

\newcommand{\X}{\ensuremath{\mathcal{X}}}
\newcommand{\Y}{\ensuremath{\mathcal{Y}}}
\newcommand{\lr}{\ensuremath{l}}
\newcommand{\Lr}{\ensuremath{L}}
\newcommand{\cl}{\ensuremath{c_l}}
\newcommand{\al}{\ensuremath{a_l}}
\newcommand{\sigl}{\ensuremath{\sigma_l}}
\newcommand{\clk}{\ensuremath{c_l^k}}
\newcommand{\alk}{\ensuremath{a_l^k}}
\newcommand{\siglk}{\ensuremath{\sigma_l^k}}
\newcommand{\sigk}{\ensuremath{\sigma^k}}
\newcommand{\fk}{\ensuremath{f^k}}
\newcommand{\Ik}{\ensuremath{I^k}}
\newcommand{\Ek}{\ensuremath{{\mathcal{E}^k}}}
\newcommand{\neighdir}{\ensuremath{\overline{n}}}
\newcommand{\neigh}{\ensuremath{\mathcal{N}}}
\newcommand{\itr}{\ensuremath{r}}
\newcommand{\Itr}{\ensuremath{R}}
\newcommand{\cdom}{\ensuremath{B}}
\newcommand{\sigdom}{\ensuremath{\Lambda}}

\begin{document}

\title{Out-of-sample generalizations for supervised manifold learning for classification} 

\author{Elif Vural and Christine Guillemot \thanks{Elif Vural and Christine Guillemot are with Centre de recherche INRIA Rennes - Bretagne Atlantique, France ({elif.vural@inria.fr}, {christine.guillemot@inria.fr}).}}


\maketitle

\begin{abstract}
Supervised manifold learning methods for data classification map data samples residing in a high-dimensional ambient space to a lower-dimensional domain in a structure-preserving way, while enhancing the separation between different classes in the learned embedding. Most nonlinear supervised manifold learning methods compute the embedding of the manifolds only at the initially available training points, while the generalization of the embedding to novel points, known as the out-of-sample extension problem in manifold learning, becomes especially important in classification applications. In this work, we propose a semi-supervised method for building an interpolation function that provides an out-of-sample extension for general supervised manifold learning algorithms studied in the context of classification. The proposed algorithm computes a radial basis function (RBF) interpolator that minimizes an objective function consisting of the total embedding error of unlabeled test samples, defined as their distance to the embeddings of the manifolds of their own class, as well as a regularization term that controls the smoothness of the interpolation function in a direction-dependent way. The class labels of test data and the interpolation function parameters are estimated jointly with a progressive procedure. Experimental results on face and object images demonstrate the potential of the proposed out-of-sample extension algorithm for the classification of manifold-modeled data sets.
\end{abstract}


\IEEEpeerreviewmaketitle

\section{Introduction}
\label{sec:intro}

\IEEEPARstart{T}{he}  recovery of low-dimensional structures in data sets not only allows understanding the data but also provides useful representations for their treatment in several problems. Data classification is among the applications that benefit from the identification of low-dimensional structures in data. Unlike unsupervised manifold learning methods such as \cite{Tenenbaum00}, \cite{Roweis00}, \cite{Belkin03}, which only take the geometric structure of data samples into account when learning a low-dimensional embedding, many recent supervised manifold learning methods  seek a representation that not only preserves the manifold structure in each class, but also enhances the separation between different class-representative manifolds in the learned embedding. Meanwhile, an important problem in data classification with supervised manifold learning is the generalization of the learned embedding to novel data samples. In this work, we address the problem of constructing a continuous mapping between the high-dimensional original data space and the low-dimensional space of embedding for data classification applications.

Supervised manifold learning methods can be categorized into two groups as linear and nonlinear algorithms. Linear methods such as  \cite{YanD07}, \cite{Hua12}, \cite{Yang11}, \cite{Zhang12}, and \cite{GaoMZGL13} learn a linear projection that maps data into a lower-dimensional space such that the proximity of neighboring samples from the same class is preserved, while the distance between samples from different classes is increased. Nonlinear methods such as \cite{Raducanu12} have a similar classification-driven objective, while the new coordinates of data samples in the low-dimensional space are computed with a nonlinear learning process based on a graph representation of data. As linear dimensionality reduction methods compute a linear projection, they have the advantage that the generalization of the embedding for initially unavailable data samples is immediate and given by the learned linear operator. However, with linear methods samples from different classes are not linearly separable in the learned embedding, unless they are already linearly separable in the original high-dimensional space, which is rarely the case. The separation between different classes is an important factor that influences the performance of classification. Nonlinear dimensionality reduction methods achieve a better separation as a result of their relative flexibility in learning the coordinates. In fact, nonlinear methods such as \cite{Raducanu12}, or nonlinear adaptations of the above linear methods, typically learn data representations where different classes become even linearly separable. However, one difficulty of using nonlinear methods is that they compute an embedding only in a pointwise manner, i.e., data coordinates in the low-dimensional domain are computed only for the initially available training data and are not generalizable to the test data in a straightforward way. Hence, an important issue that needs to be addressed in order to benefit from nonlinear manifold learning methods in classification is the generalization of the embedding to novel data samples. 

The generalization of the learned embedding to new samples is referred to as the out-of-sample extension problem in manifold learning. Several previous works have addressed the out-of-sample problem. The study in \cite{Bengio04} focuses on the extension of manifold learning methods that compute data coordinates in the form of the eigenvectors of a data kernel matrix. It is shown that in such a setting the Nystr\"om method can be used to compute eigenfunctions that coincide with the eigenvectors on the training samples and generalize them to the continuous domain. In fact, the out-of-sample extension with the Nystr\"om formula as proposed in \cite{Bengio04} can also be derived from the kernel ridge regression framework, by removing the regularization term and imposing the constraint that the data coordinates of training samples be given by the eigenvectors of the data kernel matrix \cite{ChenWG13}. Next, several out-of-sample extension algorithms rely on the construction of an interpolation function between the high- and low-dimensional domains. Some families of interpolation functions used in manifold learning extensions are polynomials \cite{QiaoZWZ13}, sparse linear combinations of functions in a reproducing kernel Hilbert space (RKHS) \cite{ChenWG13}, and sparse grid functions \cite{PeherstorferPB11}. In \cite{StrangeZ11}, the out-of-sample extension of general manifold learning methods is achieved by computing a local projection of the high-dimensional space to the low-dimensional domain with a similarity transformation of the local PCA bases. There are also some extension methods designed for particular manifold learning algorithms. The study in \cite{Trosset08} proposes an out-of-sample generalization of the multidimensional scaling (MDS) method, which is based on an interpretation of MDS as a least squares problem. Similarly, the method proposed in \cite{ChinS08} presents a generalization for maximum variance unfolding \cite{WeinbergerS06}. 

Meanwhile, all of the above methods address the out-of-sample extension problem in an unsupervised setting, i.e., no class label information of input data samples is used. In a classification problem, on the other hand, different classes are often assumed to lie on different manifolds, e.g., in a face recognition problem, the face images of each individual form a different manifold, and supervised manifold learning methods map these class-representative manifolds to different manifolds in the low-dimensional domain. Therefore, class labels of data samples and the fact that different classes are concentrated around different low-dimensional structures should be taken into account when constructing an out-of-sample extension for classification applications. Besides this, many of the above unsupervised extension methods are even not applicable in the supervised setting. For instance, the popular Nystr\"om extension \cite{Bengio04} considers embeddings given by the eigenvectors of a symmetric kernel matrix. Then, in order to embed a novel point, the kernel is evaluated between the novel point and each training point. Meanwhile, in supervised manifold learning, the value of the kernel depends not only on data sample pairs but also on the class labels of the samples. The kernel usually takes positive values for sample pairs from the same class and negative values for those from different classes, e.g., as in \cite{Raducanu12}. Hence, the Nystr\"om method does not have a straightforward generalization for supervised manifold learning.

In this paper, we propose a method for constructing out-of-sample generalizations of supervised manifold learning algorithms for classification. In order to extend the embedding (learned with any supervised algorithm) to novel points, we compute a radial basis function (RBF) interpolation function from the high-dimensional space to the low-dimensional one. We optimize the parameters of the interpolation function such that it maps initially unavailable test points as close as possible to the embeddings of the manifolds of their own class in the low-dimensional domain. This  is achieved with a progressive estimation of the class labels of test points while gradually updating the parameters of the interpolation function at the same time. As the proposed method makes use of test points in the construction of the interpolation function, it can be considered as a semi-supervised solution for obtaining an out-of-sample extension. Another criterion that is taken into account in  the optimization of the parameters of the interpolation function is the regularity of the interpolation function. We find that the regularity of the interpolation function can be adjusted by optimizing its scale parameters to minimize a regularization objective, which controls the magnitude of the interpolation function gradient, while allowing sharp directional derivatives to occur only along the class separation boundaries in order to attain an effective separation between different classes. Experimentation on several image data sets shows that the proposed method can be effectively used in the classification of data of intrinsic low dimension. The proposed out-of-sample extension method is general and can be coupled with any supervised manifold learning algorithm. 

The rest of the paper is organized as follows. In Section \ref{sec:prob_form} we briefly overview some supervised manifold learning methods and formulate the out-of-sample extension problem. In Section \ref{sec:alg_desc} we describe the proposed method for classification-driven out-of-sample extensions for supervised manifold learning. In Section \ref{sec:discussion} we discuss some aspects of the proposed algorithm, where we analyze its complexity and interpret it within the context of regression.  In Section \ref{sec:exp_results} we present some experimental results and in Section \ref{sec:concl} we conclude.

\section{Overview of manifold learning}
\label{sec:prob_form}

\subsection{Manifold learning for classification}
\label{ssec:manif_learn_overv}

Given a set of data samples $ \{x_i \}_{i=1}^N \subset \Rn$ that reside in a high-dimensional space $\Rn$, manifold learning computes a new representation $y_i  \in \Rd$ in a lower-dimensional domain $\Rd$ for each data sample $x_i$. Manifold learning methods generally assume that the samples $ \{x_i \} $ come from a model of low intrinsic dimension and search for an embedding that significantly reduces the dimension of the data ($d \ll n$) while preserving certain geometric properties. Different methods target different objectives in computing the embedding. The ISOMAP method computes an embedding such that Euclidean distances in the low-dimensional domain are proportional to the geodesic distances in the original domain \cite{Tenenbaum00}, while LLE looks for an embedding that preserves local reconstruction weights of data samples in the original domain \cite{Roweis00}. The Laplacian eigenmaps algorithm \cite{Belkin03} first constructs a graph from the data samples where nearest neighbors are typically connected with an edge. The graph Laplacian matrix is given by $L=D-W$, where $W$ is the $N \times N$ weight matrix whose entries are usually computed based on a kernel $W_{ij}=K(x_i, x_j)$, and $D$ is a diagonal degree matrix given by $D_{ii} = \sum_j W_{ij}$. The embedding with Laplacian eigenmaps is then learned by solving  
\[
 \min_{Y \in \R^{N \times d}} \tr(Y^T L Y ) \quad \text{ s.t. } \quad  Y^T D Y =I
\]
where $I$ is the identity matrix. The solution to this problem is given by the $d$ eigenvectors corresponding to the smallest nonzero eigenvalues of the generalized eigenvalue problem $L z = \lambda D z$, where the coordinate vector $y_i$ for each data sample $x_i$ is given by the $i$-th row of $Y$. Intuitively, such an embedding seeks data coordinates that have a slow variation on the data graph, i.e., two neighboring points on the graph are mapped to nearby coordinates. There exist linear versions of the Laplacian eigenmaps method as well. The above problem is solved under the constraint that $Y$ be given by a linear projection of $X$ onto $\Rd$ in \cite{He04}, which is applied to face recognitions problems in \cite{HeYHNZ05} and \cite{CaiHHZ06}.

Recently, many extensions have been proposed for manifold learning for classification. Most of these methods are supervised adaptations of the Laplacian eigenmaps algorithm. In order to achieve a good separation between the classes, an embedding is sought where data coordinates vary slowly between neighboring samples of the same class and change rapidly between neighboring samples of different classes. The algorithm proposed in \cite{Raducanu12} formalizes this idea by defining two graphs that respectively capture the within-class and between-class neighborhoods. Denoting the weight matrices of these two graphs by $W_w$ and $W_b$, and the corresponding Laplacians by $L_w$ and $L_b$, the method seeks an embedding that solves
\begin{equation}
\label{eq:obj_supLapEmb}
 \min_{Y  \in \R^{N \times d}} \tr( Y^T L_w Y)  - \mu \, \tr( Y^T L_b Y)  \quad \text{ s.t. } \quad  Y^T D_w Y = I 
\end{equation}
where $\mu>0$. The method proposed in \cite{Wang09} employs an alternative Fisher-like formulation for the supervised manifold learning problem where the embedding is obtained by solving
\begin{equation}
\label{eq:obj_supFisherEmb}
\max_{z} \frac{z^T L_b z}{z^T L_w z}.
\end{equation}
However, the problem is solved under the constraint $z^T = v^T X$ in order to obtain a linear embedding, where $X=[x_1 \dots x_N]$ is the $n \times N$ data matrix and $v \in \R^{n \times 1}$ defines a projection. Variations over this formulation can be found in several other works such as \cite{YanD07}, \cite{Hua12}, \cite{Yang11}, \cite{Zhang12}, \cite{GaoMZGL13} and\cite{XuYTLZ07}.

\subsection{Out-of-sample extensions}
\label{ssec:overv_oos_ext}

While most manifold learning methods learn the coordinates of only initially available data samples, in many applications including classification, the generalization of the learned embedding to the whole data space is an important problem. Given a set of data samples $ \{x_i \}_{i=1}^N \subset \Rn$ in the high-dimensional ambient space and their corresponding coordinates $  \{y_i \}_{i=1}^N \subset \Rd$ in a low-dimensional space, the out-of-sample extension problem consists of constructing a mapping 
$
f: \Rn \rightarrow \Rd
$
such that $f$ gives the learned embedding $f(x_i)=y_i$ on the available data samples while generalizing the embedding to all points in $\Rn$.

A popular out-of-sample generalization algorithm is presented in \cite{Bengio04}, based on the Nystr\"om formula. This method proposes a generalization for manifold learning algorithms that compute the coordinates based on an eigenvalue problem $M y^k = \lambda_k \, y^k$, where the symmetric matrix $M$ is given by a data-dependent kernel $M_{ij}= \tilde M(x_i, x_j)$ and $y^k$ is the $k$th eigenvector of $M$, which defines the $k$th dimension of the data coordinates. The exact expression of the kernel matrix $M$ as a function of the weight matrix $W$ depends on the manifold learning algorithm to be generalized, as different algorithms target different objectives. The out-of-sample extension proposed in \cite{Bengio04} is then given by the function $f(x) = [f^1(x) \dots f^d(x)]$, where
\begin{equation}
\label{eq:Nystrom}
\fk(x)= \frac{1}{\lambda_k} \sum_{i=1}^N y^k_i \tilde M(x,x_i)
\end{equation}
and $y_i=[y^1_i \dots y^d_i]$ are the coordinates of the embedding of $x_i$ in $\Rd$. This defines a continuous function that coincides with the embedding at the initially available points $f(x_i)=y_i$. While this popular method provides straightforward generalizations of many manifold learning algorithms such as ISOMAP, LLE, and Laplacian eigenmaps, it cannot be used with most supervised manifold learning methods. The reason is that, although the data kernel matrix $M$ is assumed to be a general symmetric matrix (not necessarily positive semi-definite) in \cite{Bengio04}, the entries of this matrix in supervised methods are not only dependent on the data samples $x_i$, but also on their class labels. For instance, in \eqref{eq:obj_supLapEmb}, the kernel matrix $M$ is a normalized version of the matrix $L_w - \mu L_b$, which is determined with respect to data class labels. In this case, the Nystr\"om formula (\ref{eq:Nystrom}) cannot be applied as $\tilde M(x,x_i)$ is not priorly known for a test sample of unknown class. 

Several out-of-sample extension methods such as those based on fitting a particular type of interpolation function as in \cite{ChenWG13}, \cite{QiaoZWZ13}, and \cite{PeherstorferPB11} can be applied for generalizing supervised embeddings by fitting a function $f$ to the priorly learned $(x_i, y_i)$ pairs. However, as this gives a generalized embedding based only on an approximation objective that does not take into account the class information of data, its classification performance is likely to be suboptimal. 

In this paper, we propose to learn an interpolation function in an application-aware manner. The proposed method not only makes use of the initially available training samples $(x_i, y_i)$, but also exploits the test samples of unknown class in the learning, by jointly estimating the interpolation function parameters and the class labels of test samples. We describe this method in Section \ref{sec:alg_desc}.

\section{Out-of-sample extensions for classification}
\label{sec:alg_desc}

\subsection{Formulation of the out-of-sample problem}

We begin with a formalization of the classification-based out-of-sample extension problem. Let $\M_1, \M_2, \dots \M_M \subset \Rn$ be $M$ compact manifolds representing $M$ different classes in the original ambient space $\Rn$. Let $\E$ be an embedding of the manifolds $\{ \M_m \}$ in a lower-dimensional space $\Rd$
\[
\E: \bigcup_m {\M_m} \rightarrow \Rd
\]
such that each manifold $\M_m \subset \Rn$ is mapped to $\E(\M_m) \subset \Rd $. The restriction of $\E$ to each manifold is assumed to be continuous and the embeddings of different manifolds are assumed to be disjoint. We consider that the data samples are drawn from a probability measure $\nu$ on $\Rn$ such that the samples of each class $m$ are concentrated around the manifold $\M_m$. Let $\nu_m$ denote the probability measure of class $m$ having a support region $S_m$ in $ \Rn$, where $\M_m \subset S_m$. We denote by $P_{\M_m}(x)$ a projection of the point $x$ onto the manifold $\M_m$, which is a point on $\M_m$ of minimal distance to $x$ 
\[
\| x - P_{\M_m}(x)  \|  = \min_{x' \in \M_m} \| x - x'  \|.
\]
Here $\| \cdot \|$ denotes the usual $\ell_2$-norm in the Euclidean space.

As for the solution set of interpolation functions, we consider a compact set $\Hs$ of differentiable functions from $\Rn$ to $\R$, where $f: \Rn \rightarrow \Rd$ belongs to $\Hs^d$ given by the $d$-dimensional Cartesian product of $\Hs$. An interpolation function that is suitable for classification should map points $x$ from class $m$ as close as possible to the set $\E(\M_m)$, so that they can be correctly classified with respect to their representation in $\Rd$. We thus define the embedding error of $f$ with respect to its deviation from the embedding of the projection of a point onto the manifold of its own class. The total embedding error of a function $f $ over all classes is then given by 
\begin{equation}
\label{eq:emb_error_term}
E(f) := \sum_m \int_{S_m}  \| f(x) - \E \left( P_{\M_{m}}(x) \right)  \|^2 \, d \nu_m(x).
\vspace{-10pt}
\end{equation}

\begin{figure}[t]
 \centering
  \includegraphics[width=9cm]{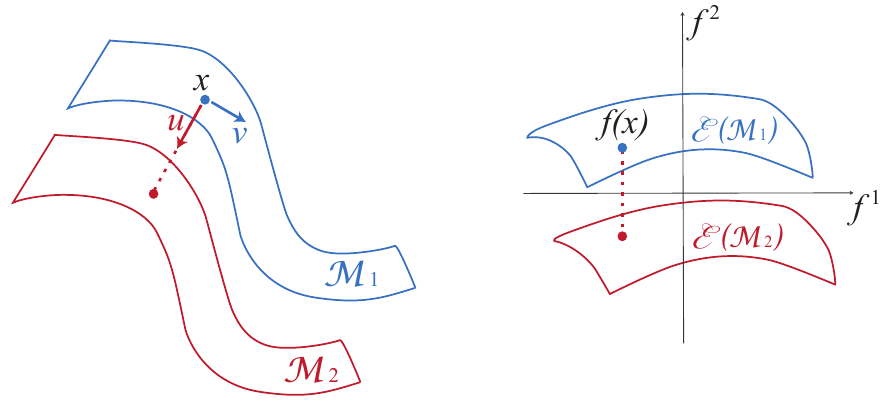}
  \caption{Illustration of supervised manifold learning and out-of-sample interpolation. Manifolds $\M_1$ and $\M_2$ representing two different classes are embedded in a lower-dimensional domain such that they are separable along dimension $k=2$, but not along dimension $k=1$. The second component $f^2(x)$ of the interpolation function $f(x)= [f^1(x) \, f^2(x) \dots f^k(x)]$ should then have a sufficiently strong directional derivative $\nabla_u f^2(x)$ along direction $u$ at $x$, in order to reinforce the separation achieved by the supervised embedding, while it should vary smoothy along direction $v$. Meanwhile, the first component $f^1(x)$ of the interpolation function should have a slow variation along both directions $u$ and $v$ as the embeddings $\E(\M_1)$ and $\E(\M_2)$ are not separable along dimension $k=1$.}
  \label{fig:illlus_embed}
  \vspace{-17pt}
\end{figure}

The distributions $\nu_m$ are usually not explicitly known in practice. In order to avoid overfitting to training data, it is useful to enforce some regularity properties for the interpolation function $f$. A smoothness constraint can be imposed by controlling the total gradient magnitude. Meanwhile, nonlinear supervised manifold learning methods, whose extensions are targeted in this paper, typically learn a representation where different classes are likely to become linearly separable in the learned embedding.  The coordinates defining the embedding are orthogonal when given by the eigenvectors of a symmetric kernel, or ``nearly orthogonal'' when given by the generalized eigenvalue problem \eqref{eq:obj_supLapEmb}. Different groups of classes are then expected to become separable along different dimensions of the learned embedding, which is also easy to confirm experimentally (see, e.g., Figure \ref{fig:embedLap}). Thus, when learning an interpolation function $f$, in order to enhance the separation between different classes, it is desirable to have sufficiently strong derivatives along the directions defining the boundaries of the distributions of different classes in the ambient space, especially for the components $\fk$ of $f$ for which the considered classes are separable at dimension $k$ of $\Rd$. This is illustrated in Figure \ref{fig:illlus_embed}. Given a dimension $k \in \{1, \dots, d \}$, let 
\begin{equation*}
\begin{split}
\Ik=\{ (m,p): &\max \Ek(\M_m) < \min \Ek(\M_p) \\
\text{ or } 
    &\max \Ek(\M_p) < \min \Ek(\M_m)  \}
\end{split}
\end{equation*}
denote the set of indices of manifold pairs whose embeddings are separable at dimension $k$, where $\Ek(\M_m)$ denotes the $k$th dimension of the embedding  $\E(\M_m)$.
Let $\nabla_v \fk$ denote the directional derivative of $\fk$ along the direction $v$. For a point $x$ from class $m$, let 
$
u_p(x) := (x - P_{\M_p}(x)) /  \| x - P_{\M_p}(x) \| 
$
denote the unit vector corresponding to the direction of projection of $x$ onto the manifold $\M_p$ of class $p$, where $p \neq m$. Then, we would like to learn an interpolation function $f$ such that the directions along which $\fk$ has the strongest derivatives coincide with the directions of the projections of points onto the manifolds of other classes. The total derivative magnitude along the directions of projection onto other classes, normalized by the average derivative magnitude is given by
\begin{equation}
\label{eq:deriv_separ_dir}
D(\fk):=\sum_{(m,p) \in \Ik}   \int_{S_m}
  \frac{ \left \| \nabla_{u_p(x)} \, \fk (x) \right \|  }
  { \Ev_v  \|  \nabla_{v} \, \fk (x)   \|  }
  \, d \nu_m(x)
\vspace{-10pt}
\end{equation}
where $\fk$ is assumed to have nowhere vanishing gradient and $ \Ev_v \|  \nabla_{v} \, \fk (x)   \| $ denotes the mean directional derivative magnitude, induced from the overall distribution of data over all classes.\footnote{
The mean directional derivative can be formally defined as follows. Let $p_\nu$ denote the probability density function corresponding to the probability measure $\nu$, and let $S_t(x) \subset \R^n$ denote the $(n-1)$-dimensional sphere of radius $t$ centered at $x$.  The expected value of the directional derivative of $\fk$ at $x$ is then given by
\[
 \Ev_v  \|  \nabla_{v} \, \fk (x)   \| = \lim_{t \rightarrow 0} 
 \frac{\int_{S_t(x)} p_\nu(x+tv) \, \|  \nabla_{v} \, \fk (x)   \| dS  }
 {\int_{S_t(x)} p_\nu(x+tv) \, dS }
\]
where $v$ denotes the unit surface normal in the direction of the surface element $dS$.
}
The normalization of the directional derivative by the average derivative aims to measure the derivative magnitude along separation boundaries relatively to the mean derivative magnitude.

While the presence of sufficiently strong directional derivatives along separation boundaries is expected to enhance the separation between classes with the learned function, it is useful to control the smoothness of the interpolation function by preventing it from attaining arbitrarily high gradient magnitudes. We thus define the total gradient magnitude 
\begin{equation}
\label{eq:tikhonov_grad}
G(\fk):=\sum_m \int_{S_m} \frac{ \| \nabla \fk(x) \|  }{ \Ev_v  \|  \nabla_{v} \, \fk (x)   \| }  \, d \nu_m(x)
\end{equation}
which is also normalized by the average directional derivative so that it is comparable to the term in \eqref{eq:deriv_separ_dir}.

From \eqref{eq:deriv_separ_dir} and \eqref{eq:tikhonov_grad},  one can define an overall regularization objective $R$ that increases with the total gradient magnitude $G$ and decreases with the directional derivative magnitude along separation boundaries $D$. One way to define the regularization objective $R$ is as
\begin{equation*}
R(f)= \sum_{k = 1}^d \left( G(\fk) - \lambda D(\fk) \right)
\vspace{-10pt}
\end{equation*}
where $\lambda > 0$. 

Finally, combining the embedding error in \eqref{eq:emb_error_term} and the above regularization term, we formulate the search of the interpolation function $f$ as the optimization of the following problem
\begin{equation*}
\begin{split}
f = \arg \min_{h \in \Hs^k}  \,   
  & \big(  \sum_m \int_{S_m}  \| h(x) - \E \left( P_{\M_{m}}(x) \right)  \|^2 \, d \nu_m(x)  \\
  &+ \alpha R(h)   \big)
\end{split}
\end{equation*}
where $\alpha >0$. A solution to the above problem exists as $\Hs^k$ is compact and the objective function is continuous. 

In a real setting, the distributions $\nu_m$ of data are often not explicitly known and one has access to a set of samples $\X=\{ x_i  \}_{i=1}^Q$ drawn from these distributions. Let $C_i \in \{ 1, \dots, M \}$ denote the class label of the data sample $x_i$, and $\neigh(x_i)$ be the set of nearest neighbors of $x_i$ in $\X$ (which can be chosen for instance as the $K$-nearest neighbors of $x_i$ with respect to the Euclidean distance in $\Rn$). Let 
\begin{equation*}
\neighdir(x_i) =  \left \{ \frac{x_i - x_j}{\| x_i - x_j \|  }: x_j \in \neigh(x_i) \right \}
\end{equation*}
denote the set of unit vectors that indicate the directions of the neighbors of $x_i$, and 
\begin{equation*}
\neighdir_p(x_i) =  \left \{ \frac{x_i - x_j}{\| x_i - x_j \|  }: x_j \in \neigh(x_i) , C_j = p \right \}
\end{equation*}
denote the set of unit directions given by the nearest neighbors of $x_i$ within class $p$. We can then define the empirical embedding error $\hat E(f)$ as 
\begin{equation*}
 \hat E(f) = \sum_m \sum_{i: \, C_i = m}  \| f(x_i) - \E \left( P_{\M_{m}}(x_i) \right)  \|^2 
\end{equation*}
and the empirical counterpart $\hat R(f)$ of the regularization term $R(f)$ as
\begin{equation}
\label{eq:reg_obj_disc}
\hat R (f) = \sum_{k = 1}^d \left( \hat G(\fk) - \lambda \hat D(\fk) \right)
\end{equation}
where 
\begin{equation}
\label{eq:regobj_disc_G}
\hat G(\fk) :=   \sum_{i} \frac{ \| \nabla \fk(x_i) \|  }{  |  \neighdir(x_i)  |^{-1}  \sum_{v \in \neighdir(x_i) }   \| \nabla_{v} \, \fk (x_i)  \| }
\end{equation} 
\begin{equation}
\label{eq:regobj_disc_D}
\begin{split}
\hat D(\fk):= &\sum_{(m,p) \in \Ik}  \, \,  \sum_{i: \, C_i = m}  \, \, \,
	  \frac{1}{ | \neighdir_p(x_i) | }   \\ 
	  &  \sum_{u \in {\neighdir_p(x_i) }}
  	  \frac{ \left \| \nabla_{u} \, \fk (x_i) \right \|  }
  {  |  \neighdir(x_i)  |^{-1}  \sum_{v \in \neighdir(x_i) }  \| \nabla_{v} \, \fk (x_i)  \|   }.
\end{split}
\end{equation}
In the above expressions, $| \cdot |$ denotes the cardinality of a set,  and the mean directional derivative  $\Ev_v  \|  \nabla_{v} \, \fk (x_i)   \| $ is approximated by the average derivative of $\fk(x_i)$ along the directions of the neighbors $\neighdir(x_i)$ of $x_i$. In the definition of $\hat D (\fk)$, we approximate the derivative $  \nabla_{u_p(x_i)} \, \fk (x_i) $ along the direction of the projection of $x_i$ onto $\M_p$ with the average derivative along the directions of the nearest neighbors of $x_i$ within class $p$,  where $\neighdir_p(x_i) $ is assumed to be non-empty for all $x_i $ and $p$.

Now, having defined the objective function in the empirical setting, we come back to the actual manifold learning problem. In practice, the manifolds $\M_m$ are usually not explicitly known, and manifold learning methods compute an embedding for only the initially available training samples. Let us denote by $\X_T = \{ x_i \}_{i=1}^N \subset  \X$ the set of training samples with known class labels (where $N \leq Q$), for which an embedding  $\Y_T=\{ y_i  \}_{i=1}^N$ is priorly computed with a supervised manifold learning algorithm. We assume that there exist embeddings $\E(\M_m)$ of the manifolds $\M_m$ such that the embeddings $\{ y_i \}$ of the samples of each class $m$ are concentrated around $\E(\M_m)$. Although the training samples available in practice are not guaranteed to lie exactly on a manifold in general (due to noise, imprecise measurements, or several sources of deviation from the assumed model), we make the following approximations for a sample $x_i \in \X_T$ of class $m$ for the simplicity of computations:
\begin{equation*}
P_{\M_m}(x_i) \approx x_i, 
\qquad \qquad
\E(P_{\M_m}(x_i)) \approx y_i.
\end{equation*}
The embedding error $\hat E(f)$ can then be decomposed as
$
 \hat E(f) = \hat E_T(f)  +  \hat E_O(f) 
$
where
\begin{equation*}
\hat E_T(f) =  \sum_{i=1}^N   \| f(x_i) - y_i  \|^2  
\end{equation*}
is the embedding error of the training samples $\X_T$ and
\begin{equation*}
\begin{split}
\hat E_O(f) 
& = \sum_{i=N+1}^Q  
\| f(x_i) - \E \left( P_{\M_{m}}(x_i) \right)  \|^2  \\
& =   \sum_m \sum_{ \substack{     i=N+1 \\ C_i = m } } ^Q  \| f(x_i) - \E \left( P_{\M_{m}}(x_i) \right)  \|^2 
\end{split}
\end{equation*}
is the embedding error of the other samples than the training samples (test samples in $\X \setminus \X_T$).

In the generalization of an embedding, one may wish to strictly preserve the learned coordinates of the training data $f(x_i)= y_i$.  We can thus formulate the out-of-sample extension problem for supervised manifold learning as follows:
\begin{equation}
\label{eq:overall_obj_discr}
f =  \arg \min_{h \in \Hs^k}  \hat E_O(h) +  \alpha  \hat R (h)  
	\quad \text{ s.t. } \quad 
	\hat E_T(h)=0.
\end{equation}

In the above problem, if there are observations in $\X$ with unknown class labels, one needs to estimate the class labels $C_i$ for $N< i \leq Q$. In the rest the paper, we focus on this general case. In Section \ref{ssec:interp_comp}, we describe an algorithm that computes an interpolation function with a joint and progressive estimation of the function parameters and the class labels of data.

\subsection{Construction of the interpolation function}
\label{ssec:interp_comp}

In this study, we select the set $\Hs$ of interpolation functions for the out-of-sample extension problem as the radial basis functions (RBFs)
\begin{equation*}
\Hs = \left \{ g: g(x) = \sum_{\lr=1}^{\Lr} \cl \, \phi \left(  \frac{\| x -  \al \|}{\sigl} \right) \right  \}
\end{equation*}
where $\phi: \R \rightarrow \R^{+}$ is a differentiable kernel. The coefficients $\cl $, the kernel centers $\al$, and the scale parameters $\sigl $ are assumed to lie in some compact domains in $\R$, $\Rn$ and $\R^{+}$, respectively. The Gaussian function $\phi(t)=e^{-t^2}$  is a common choice for the RBF kernel due to its desirable properties such as its smoothness and rapid decay, which we also adopt in this work.

In our problem, we look for a function $f=[f^1(x) \, \dots  \, f^d(x)]: \Rn \rightarrow \Rd$ such that each dimension $\fk$ of $f$ is given by
\begin{equation}
\label{eq:int_func_form}
\fk(x) = \sum_{\lr=1}^{\Lr}  \cl^k \, \phi \left(  \frac{\| x -  \al^k \|}{\sigl^k} \right).
\end{equation}
The construction of the interpolation function $f$ is thus equivalent to the determination of the parameters $\clk$, $\alk$, $\siglk$, and the number of terms $\Lr$. 

In the optimization problem in \eqref{eq:overall_obj_discr}, the evaluation of $\hat E_O(h)$ requires the knowledge of the class labels $C_i$ of $x_i$ for $i=N+1, \dots, Q$, which are unavailable in the beginning. We propose to solve this problem with an iterative algorithm that progressively estimates the class labels and constructs a sequence of interpolation functions $f_1, \dots, f_\itr, \dots f_\Itr$ in an alternating scheme as described below. 

In iteration $\itr$ of the algorithm, we construct a function $f_\itr$ with $L_\itr$ terms. When fitting an RBF interpolation function to data, it is common practice to assign kernel centers as data points. In iteration $\itr$, we select the kernel centers $\alk = x_{\itr_l}$ as a subset of data samples $\{ x_{\itr_l} \}_{\lr=1}^{\Lr_\itr} \subset \X$, where the index sequence $\{\itr_l\}_{\lr=1}^{\Lr_\itr}$ depends on the iteration $\itr$ and denotes the indices of the data samples $\{ x_i\}$ chosen as kernel centers. Throughout the iterations, the number of terms $L_\itr$ is increased gradually such that $N=\Lr_1 < \Lr_2 < ... < \Lr_\Itr = Q$. Once the kernel centers $\alk$ are fixed, the interpolation function $f_\itr$ in iteration $\itr$, characterized by the coefficients $\{ \clk \}$ and the scale parameters $\{ \siglk \}$, $\lr=1, \dots, \Lr_\itr$, $k=1, \dots, d$, is obtained by solving the problem
\begin{equation}
\label{eq:opt_f_iter_r}
 \min_{\{ \clk  \} \subset  \cdom , \,  \{ \siglk \}  \subset \sigdom }  \hat E^\itr_O(f) +  \alpha \hat R(f)  
	\quad \text{ s.t. } \quad 
	\hat E_T(f)=0
\end{equation}
where $\cdom \subset \R$ and $\sigdom \subset \R^{+}$ are compact parameter domains (sufficiently large so that the constraint $\hat E_T(f)=0$ can be satisfied) and
\begin{equation}
\label{eq:emb_error_iter_r}
\hat E^\itr_O(f) = \sum_m \sum_{ \substack{     l=N+1 \\ C_{\itr_l} = m } } ^{\Lr_\itr}  \| f(x_{\itr_l}) - \E \left( P_{\M_{m}}(x_{\itr_l}) \right)  \|^2.
\end{equation}
The problem \eqref{eq:opt_f_iter_r} has a solution as a continuous function over a compact domain attains its minimum.

%

Before discussing the solution of \eqref{eq:opt_f_iter_r}, we first give an overview of the method. In iteration $\itr$, once the interpolation function $f_\itr$ is computed by solving \eqref{eq:opt_f_iter_r}, we estimate the class label of each point $x_i$ for $N+1 \leq i \leq Q$ by assigning it the class label of the training point $x_j$ such that $f_\itr(x_j)$ is the closest to $f_\itr(x_i)$ in the low-dimensional domain $\Rd$:
\begin{equation}
\label{eq:NNclass_rule}
C_i = C_j : \quad j=\arg \min_q \| f_\itr(x_q) - f_\itr(x_i)  \|, \, \, 1\leq q \leq N. 
\end{equation}
At the same time, a confidence score $\mu_i$ is assigned to each estimate $C_i$ by comparing the distance of $x_i$ to its nearest neighbor $x_j$ within all classes and to its nearest neighbor $x_{j'}$ among the classes other than $C_j$:
\begin{equation}
\label{eq:confid_meas_defn}
\begin{split}
\mu_i & = \frac{\| f(x_{j'}) - f(x_i) \|}{\| f(x_j) - f(x_i) \|}: \\
 j'& =\arg \min_q \| f_\itr(x_q) - f_\itr(x_i)  \|, \, \, 1\leq q \leq N, C_{n} \neq C_{j}. 
\end{split}
\end{equation}
The confidence score $\mu_i$ thus decreases with the ``ambiguity'' in assigning $x_i$ the class label $C_i$ with respect to the nearest-neighbor decision rule in $\Rd$ via $f_\itr$.

The confidence scores $\mu_i$ obtained in an iteration are then used in the next iteration for the selection of the kernel centers. In iteration $\itr$, the kernel centers are determined based on the confidence scores computed in the previous iteration $\itr-1$ as follows. The first $N$ kernel centers $\{ \alk \}_{\lr=1}^N= \{ x_{\itr_\lr} \}_{\lr=1}^N$ consist of the training set $\X_T$, i.e., $\itr_\lr=l$ for $\lr=1, \dots, N$. The remaining kernel centers $\{ \alk \}_{\lr=N+1}^{\Lr_\itr}$ are then set as the first $\Lr_{\itr}-N$ points in $\X \setminus \X_T$ of highest confidence scores. The alternating stages of computing $f_\itr$ and estimating the class labels $C_i$ and obtaining the confidence scores $\mu_i$ are repeated until the last iteration $R$, where all data samples are included in the set of kernel centers $\{ \alk \}_{\lr=1}^Q= \X$. The interpolation function $f$ is then given by $f_\Itr$, and the class labels of the points in $\X$ are obtained by estimating them with the final interpolation function with respect to \eqref{eq:NNclass_rule}.

We now discuss the solution of the problem \eqref{eq:opt_f_iter_r}. First, observe that for any $\Lr_\itr$ input data pairs $(x_i, y_i) \in \Rn \times \Rd$ and any choice of the scale parameters $\siglk$, one can find  interpolation functions $f^k$ of $\Lr=\Lr_\itr$ terms that satisfy $f(x_i)=y_i$ as follows. 
Setting $\alk = x_l $ for $l=1, \dots, \Lr_\itr$, the constraints $\fk(x_i)=y_i^k$ yield the linear system 
\begin{equation}
\label{eq:RBFlinsys} 
\Phi^k c^k =  y^k 
\end{equation}
where $c^k=[c_1^k \dots c_{\Lr_\itr}^k]^T$ is the coefficient vector, $y^k=[y_1^k \dots y_{\Lr_\itr}^k]^T$ consists of the $k$th dimensions of $\{y_i\}$, and
\begin{equation}
\label{eq:defn_Phi_matrix}
\Phi^k_{il}= \phi \left(  \frac{\| x_i -  x_l \|}{\siglk} \right)
\end{equation}
is the matrix of RBFs evaluated at data points $x_i$. The square matrix $\Phi^k$ is  invertible if the points $x_i$ are distinct and $\phi$ is chosen as the Gaussian kernel \cite{Buhmann03}. The system \eqref{eq:RBFlinsys} then has a unique solution $c^k = (\Phi^k)^{-1} y^k$, which satisfies $\fk(x_i)=y_i^k$. 

In iteration $\itr=1$, we have $\Lr_1=N$ and all kernel centers are training points. In this case the embedding error in \eqref{eq:emb_error_iter_r} is $\hat E^1_O(f)=0$, and the optimization problem is reduced to 
\begin{equation*}
 \min_{\{ \clk  \} \subset \cdom,  \, \{ \siglk  \}  \subset \sigdom }  \hat R(f)  
	\quad \text{ s.t. } \quad 
	\hat E_T(f)=0.
\end{equation*}
Due to the above discussion, the constraint $\hat E_T(f)=0$ can be satisfied for any choice of scale parameters $\siglk$ by setting the coefficients as $c^k = (\Phi^k)^{-1} y^k$. This reduces the problem to the minimization of the regularization term $\hat R(f)$ by optimizing the scale parameters $\{ \siglk \} $ under the constraint $c^k = (\Phi^k)^{-1} y^k$
\begin{equation}
\label{eq:opt_reg_iter1}
 \min_{  \substack{     \{ \siglk \} \subset \sigdom \\ c^k = (\Phi^k)^{-1} y^k }  }  \hat R(f)  
\end{equation}
where the summations in the terms \eqref{eq:regobj_disc_G} and \eqref{eq:regobj_disc_D} of  $\hat R(f)$ run over the set of training samples $\X_T$. The regularization term is a non-convex function of the scale parameters $\{ \siglk \}$ with numerous extrema.  Meanwhile, we have experimentally observed that the variation of $\hat R(f)$ with $\sigk$ is quite regular when all scale parameters $\siglk$, $\lr = 1, \dots, \Lr_1$, in each dimension $k$ are set to a common value $\sigk$. Moreover, setting all scale parameters to the same value across each dimension also simplifies the optimization problem, as it reduces  the number of optimization variables from $\Lr_1 k $ to $k$. We thus propose to solve the problem \eqref{eq:opt_reg_iter1}  under the constraint $\siglk=\sigk $ for $\lr=1, \dots, \Lr_1$. Since the form of $\hat R(f)$ in \eqref{eq:reg_obj_disc} is decomposable into its components in different dimensions, the scale parameter of dimension $k$ is given by 
\begin{equation*}
\min_{ \substack{  \sigk \in \sigdom \\   c^k = (\Phi^k)^{-1} y^k} } \left( \hat G(\fk) - \lambda \hat D(\fk) \right).
\end{equation*}

\begin{figure}[t]
\begin{center}
     \subfigure[]
       {\label{fig:manifR2}\includegraphics[height=2.1cm]{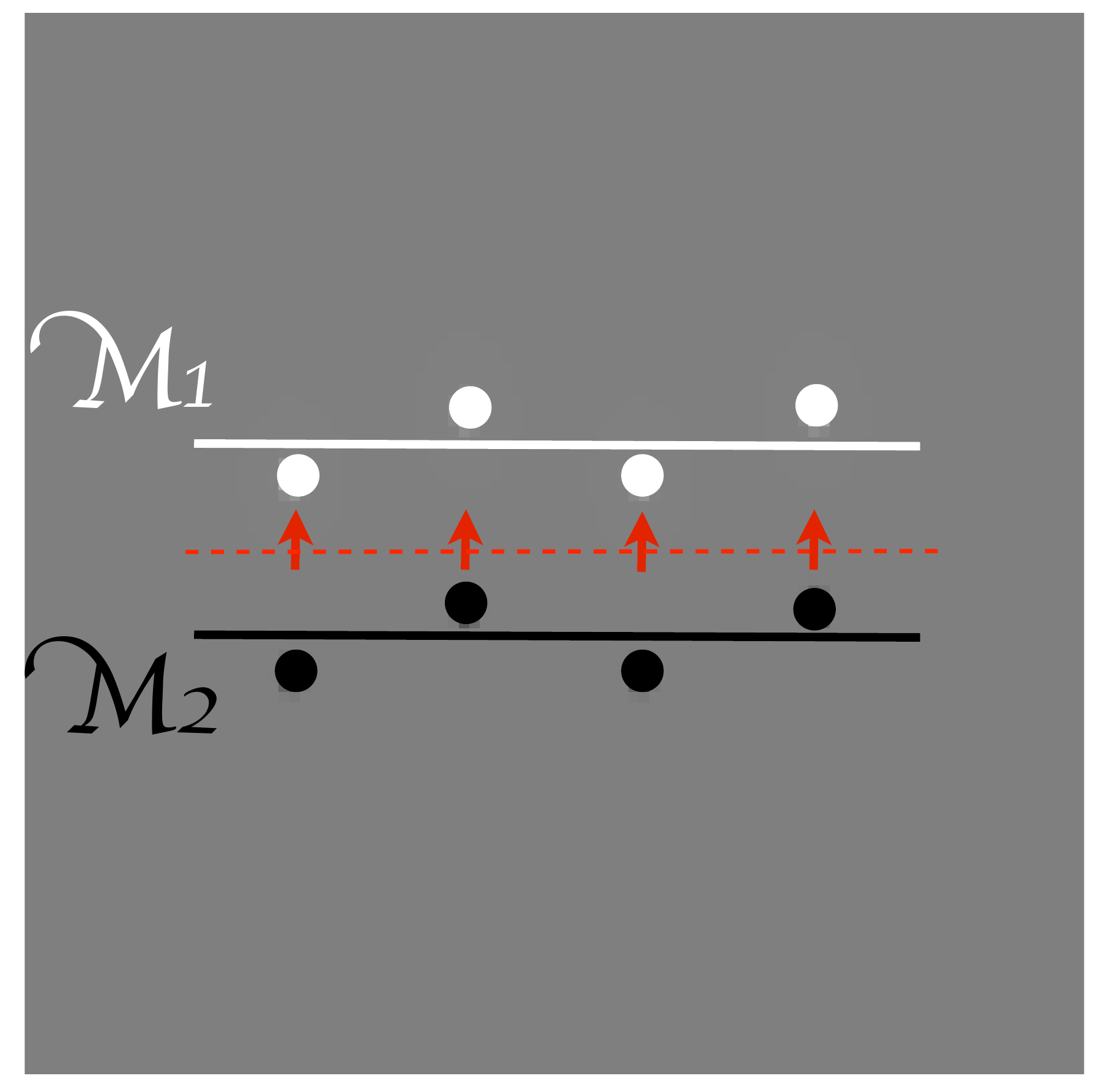}}
     \subfigure[]
       {\label{fig:embed_sigma0p5}\includegraphics[height=2.1cm]{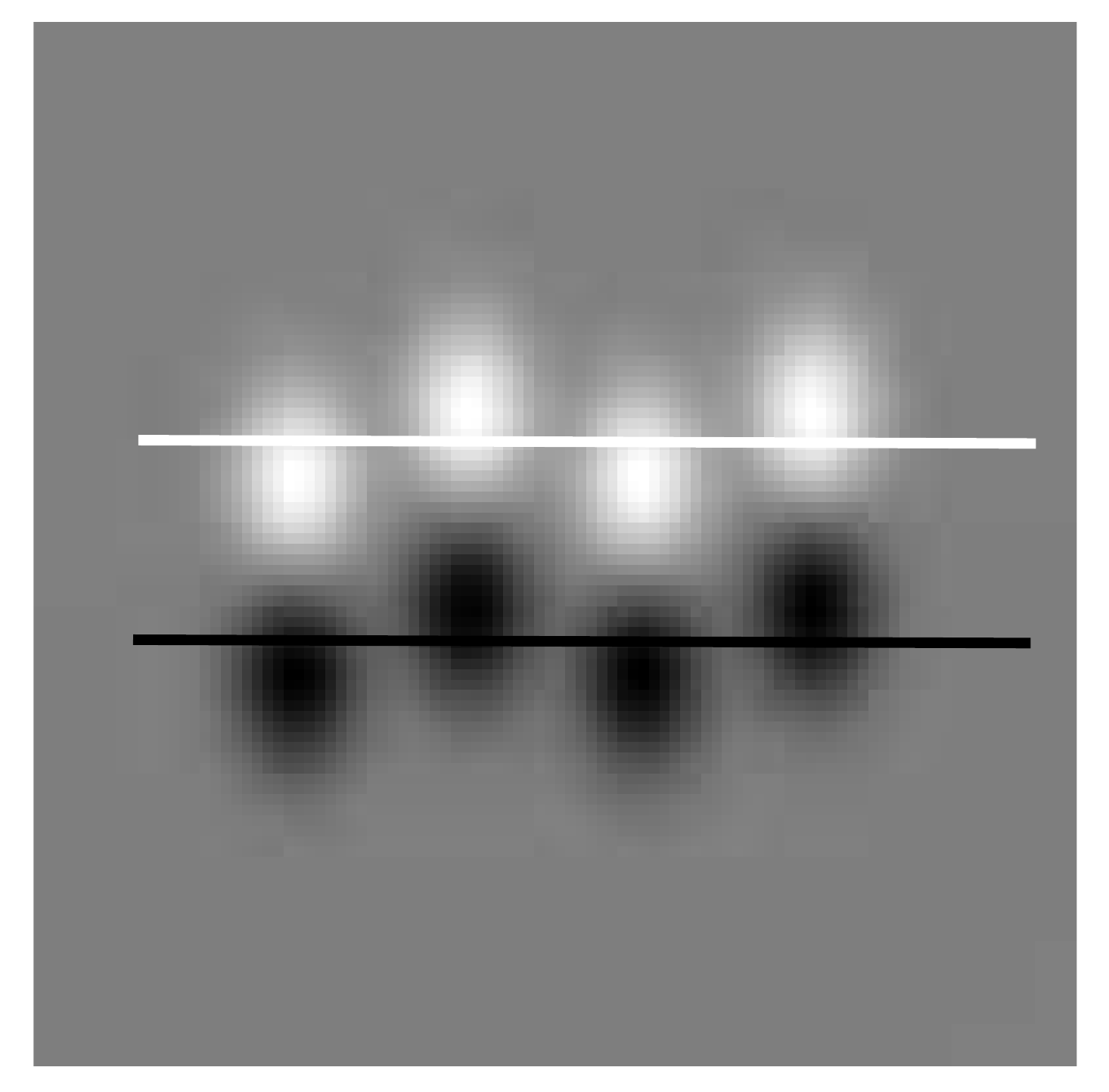}}
      \subfigure[]
       {\label{fig:embed_sigma2}\includegraphics[height=2.1cm]{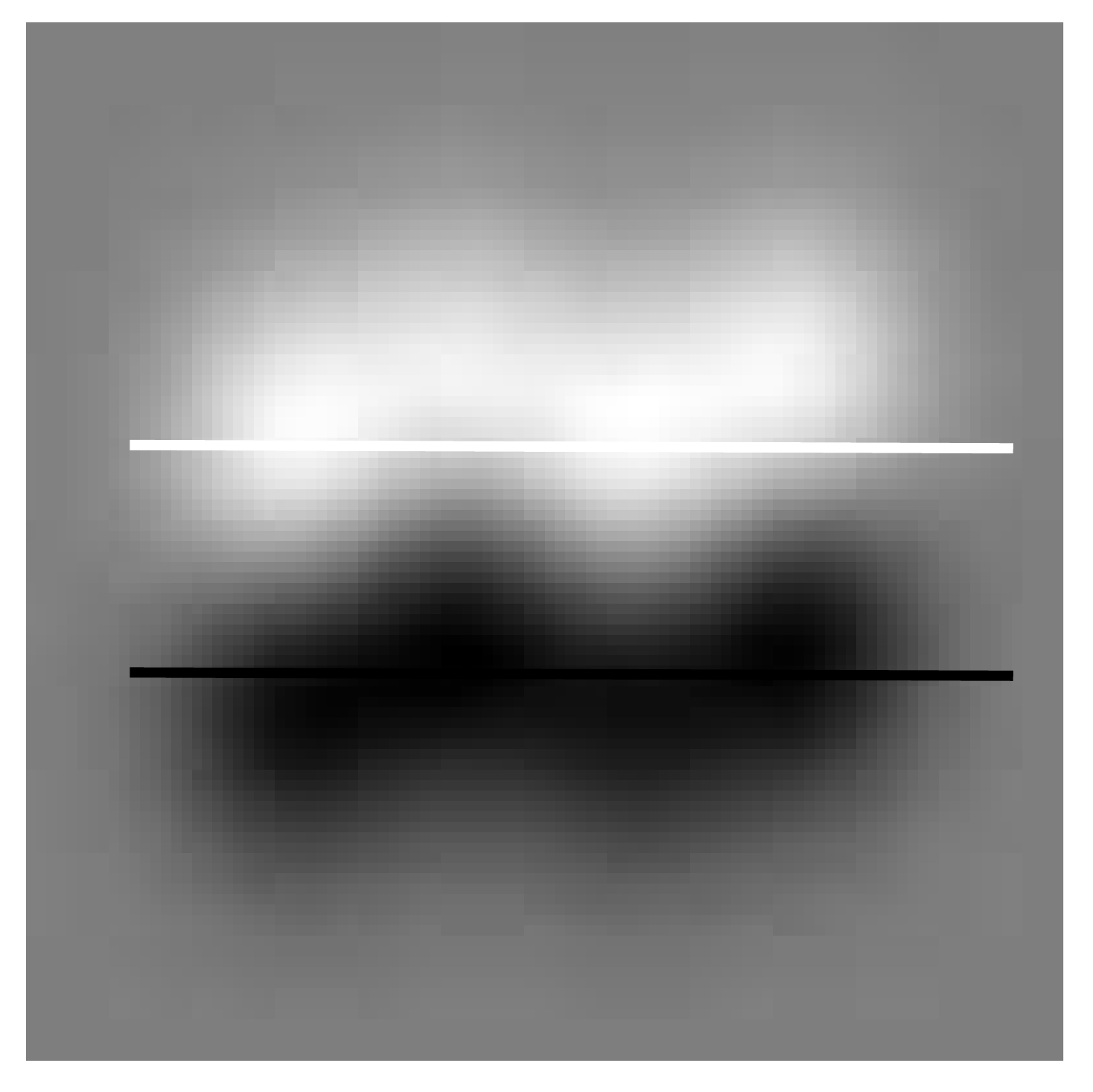}}
       \subfigure[]
       {\label{fig:embed_sigma6}\includegraphics[height=2.1cm]{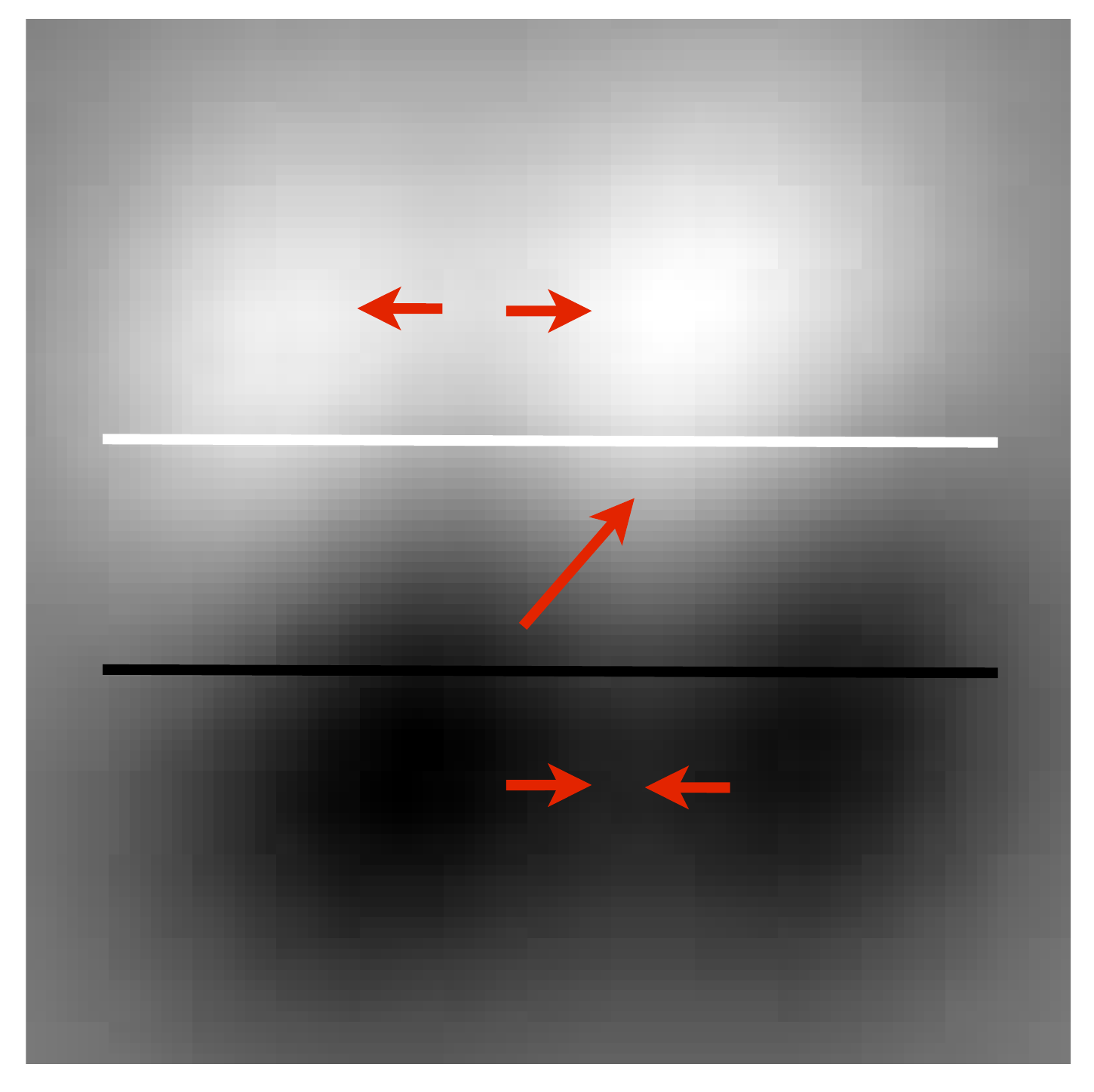}}
 \end{center}
 \caption{Illustration of the effect of the scale parameter on the accuracy of the interpolation function. (a) Manifolds $\M_1, \M_2 \subset \R^2$ representing two different classes and samples chosen from each class. An ideal interpolation function $f$ separating the two classes well in $\R$ should have gradients in the directions shown in red. (b) Function $f$ constructed with $\sigma=0.5$. The scale parameter is observed to be too small as the support of the function does not cover the manifolds well. (c) Choosing the scale as $\sigma=2$ yields a good interpolation function. (d) Choosing a too large scale parameter $ \sigma=6$ results in an overfitting of the interpolation function, with large derivatives in the indicated directions.}
 \label{fig:illus_manifR2}
 \vspace{-20pt}
\end{figure}

It is difficult to analyze the above function theoretically. Meanwhile, in practice we have observed that $\hat G(\fk) $ increases with $\sigk$ monotonically. Moreover, if the underlying embedding obtained with supervised manifold learning provides a ``balanced'' distribution of the classes across different dimensions while ensuring a sufficient separation, the total directional derivative along class separation boundaries $\hat D(\fk)$ first increases at a fast rate with $\sigk$ at small scale values, and then it stagnates or the rate of increase is highly reduced. This is due to the fact that, when the scale parameters are too small, the interpolation function is too localized around kernel centers and its support does not cover well the whole space. Then, it does not have sufficiently strong derivatives along class separation boundaries. As $\sigk$ increases, there typically exists a range for $\sigk$ where the directional derivatives along class separation boundaries are relatively stronger than those along other directions, thanks to the underlying learned embedding that separates different classes and guides the interpolation function via the condition $\fk(x_i)=y_i^k$ imposed on training samples. This range for the scale parameters coincides in general with the interval of scale parameters where a good classification performance is attained. If the scale parameters are increased beyond this range, the gradient of the function $\fk$ increases too much, resulting in an overfitting of the interpolation function, where the advantage of having sufficiently strong directional derivatives along class separation boundaries is lost as strong derivatives appear in other directions as well due to overfitting. This is illustrated with a simple example in Figure \ref{fig:illus_manifR2}. Figure \ref{fig:manifR2} shows two manifolds $\M_1, \M_2 \subset \R^2$ representing two different classes, and four training samples selected from the distribution concentrated around each manifold. Let us consider a one-dimensional embedding of the manifold samples in $\R$ such that samples from $\M_1$ and $\M_2$ are mapped respectively to $1$ and $-1$. An ideal interpolation function $f(x): \R^2 \rightarrow \R$ separating the two classes well in $\R$ should have gradients in the directions shown in red in Figure \ref{fig:manifR2}, which are orthogonal to the class separation boundary. In Figures \ref{fig:embed_sigma0p5}-\ref{fig:embed_sigma6}, an RBF interpolation function $f$ with Gaussian kernel is fitted to the training data, and $f(x)$ is plotted over the displayed region of $\R^2$, where white and black colors correspond respectively to $1$ and $-1$. The scale parameter is chosen as $\sigma=0.5$,  $\sigma=2$, and $ \sigma=6$ respectively in Figures \ref{fig:embed_sigma0p5}- \ref{fig:embed_sigma6}. The scale parameter is observed to be too small in Figure \ref{fig:embed_sigma0p5} as the support of $f$ does not cover the manifolds sufficiently. The scale parameter in Figure \ref{fig:embed_sigma2} yields an accurate interpolation function that separates the two classes well, where the directions along which $f$ has strong derivatives are close to the directions shown in Figure \ref{fig:manifR2}. Meanwhile, the selection of a too large value for the scale parameter in Figure \ref{fig:embed_sigma6} results in an overfitting of the interpolation function. In particular, strong directional derivatives are observable  in directions other than the class separation boundary directions as well due to overfitting, e.g., the directions shown in red in Figure \ref{fig:embed_sigma6}.

In optimizing $\sigk$, we look for an interval where $\hat D(\fk)$ is large enough while $ \hat G(\fk)$ is not too high. We set the weight parameter $\lambda$ to a value where the effects of both of these terms are visible, often yielding a nonmonotonic variation of the overall regularization term $ \hat G(\fk) - \lambda \hat D(\fk) $, which first decreases with $\sigk$ due to the sharp increase in $\hat D(\fk)$ and then increases with $\sigk$ due to the stagnation of $D(\fk)$ and the continuing increase in the first term $\hat G(\fk)$. The optimal value of $\sigk$ can then be found easily with a simple descent or line search algorithm by minimizing the one-dimensional regularization term $\hat G(\fk) - \lambda \hat D(\fk)$. We finally note that other configurations of these two terms $\hat D(\fk)$ and $\hat G(\fk)$ in a regularization objective $\hat R(f)$ (rather than a linear combination) may also be possible, depending on the underlying embedding. This will be discussed in more detail in Section \ref{sec:exp_results}, as well as the links between the regularization objective $\hat R(f)$ and the classification performance.

Having examined the computation of the scale parameters and the coefficients of $f_1$ in iteration $1$, we now discuss the solution of the problem \eqref{eq:opt_f_iter_r} in a general iteration $\itr$. Due to the iterative estimation of the class labels and the calculation of the function parameters, the class labels $C_{\itr_\lr}$ of the points $x_{\itr_\lr}$ contributing to the embedding error \eqref{eq:emb_error_iter_r} are already estimated in the previous iteration. The manifolds $\M_m $ and the embedding $\E$ are not explicitly known in the term $\E \left( P_{\M_{m}}(x_{\itr_l}) \right)$. However, relying on a locally linear approximation of the manifolds, one can estimate the projection of a point $x$ onto $\M_m$ as a convex combination of its nearest neighbors, which can then be used to compute $\E \left( P_{\M_{m}}(x_{\itr_l}) \right)$.\footnote{Note that, although the interpolation function of the previous iteration gives an estimate of the embedding of a point $x$ as $f_{\itr-1}(x)$, it is more reliable to update the embedding by projecting $x$ onto the manifold $\M_m$. This is because the embedding $f_{\itr-1}(x)$ employs no priors on the class label of $x$ and is indeed used to estimate the class label of $x$, while the recomputation of the embedding as $\E \left( P_{\M_{m}}(x) \right)$ uses the estimated class label of $x$. In fact, $\E \left( P_{\M_{m}}(x) \right)$ coincides with the value of the updated interpolation function $f_{\itr}(x)$ of iteration $\itr$ for $x=x_{\itr_\lr}$ as discussed below.} Denoting the indices of the $K$ nearest neighbors of $x$ within the training samples of class $m$ as $\{ a_i\}_{i=1}^K$, and the set of nearest neighbors as $\neigh_m(x)=\{ x_{a_i} \}_{i=1}^K$, the projection is approximated as
\begin{equation*}
P_{\M_{m}}(x) \approx \sum_{i=1}^K w_i \, x_{a_i}
\end{equation*}
where $w=[w_1 \dots w_K]^T$ is the vector of weights given by
\begin{equation}
\label{eq:quad_prog_embed}
w=\arg \min_{v} \| x -  \sum_{i=1}^K v_i \, x_{a_i} \|^2
\quad \text{ s.t. } \quad
v_i \geq 0, \, \sum_{i=1} v_i =1
\end{equation}
which can be solved with quadratic programming. From the continuity assumption of the embeddings, the embedding $\E \left( P_{\M_{m}}(x) \right)$ of $ P_{\M_{m}}(x)$ is then estimated as
\begin{equation}
\label{eq:embedding_Pmx}
\E \left( P_{\M_{m}}(x) \right) \approx \sum_{i=1}^K w_i \, y_{a_i}
\end{equation}
where $y_{a_i}$ are the coordinates of $x_{a_i}$ in the learned embedding in $\Rd$. 

Letting $y_{\itr_\lr} = \E \left( P_{\M_{C_{\itr_\lr}}}(x_{\itr_\lr}) \right)$, the total embedding error is given by 
\begin{equation*}
\hat E^\itr(f) =  \hat E^\itr_O(f) + \hat E_T(f) = \sum_m \sum_{ \substack{     l=1 \\ C_{\itr_l} = m } } ^{\Lr_\itr}  \| f(x_{\itr_l}) - y_{\itr_\lr}   \|^2.
\end{equation*}
Since in iteration $\itr$ an interpolation function of $\Lr_\itr$ terms is constructed, for any choice of the scale parameters $\{ \siglk \}$, fitting the coefficients to the observations as $c^k = (\Phi^k)^{-1} y^k$ yields $\hat E^\itr(f) =0$, which immediately satisfies the constraint $\hat E_T(f)=0$ on training samples. It then remains to minimize the regularization term by optimizing the scale parameters as in \eqref{eq:opt_reg_iter1}.\footnote{In practice the optimization of scale parameters can be omitted for $r>1$ and the scale parameters can be set to the $\sigk$ values obtained in iteration $r=1$ in order to speed up the algorithm without much change in the performance, as the reoptimization of the scale parameters results in $\sigk$ values in the vicinity of those obtained at iteration $\itr=1$ in general.} This concludes the description of the proposed method. As the proposed algorithm employs unlabeled test samples in learning an out-of-sample extension, we call it Semi-supervised Out-of-Sample Interpolation (SOSI). The method is summarized in Algorithm \ref{alg:sosi}.

\begin{algorithm}[h]
\caption{Semi-supervised Out-of-Sample Interpolation (SOSI)}

\begin{algorithmic}[1]
\begin{footnotesize}

\STATE
\textbf{Input:} \\
$\X=\{ x_i \}_{i=1}^Q \subset \Rn$: Set of labeled and unlabeled data samples\\
$\{ C_i \}_{i=1}^N $: Class labels of training data $\X_T=\{ x_i \}_{i=1}^N \subset \X$, where $N < Q$.  \\

\STATE
\textbf{Initialization:} Assign number of iterations $R$ and number of RBF terms $\{ \Lr_\itr\}_{\itr=1}^\Itr$ in each iteration such that $\Lr_1=N$, $\Lr_\Itr=Q$ (possibly with equispaced intervals between $N$ and $Q$)

\FOR{$\itr=1$}

\STATE 
Set kernel centers $\alk = x_\lr$ for $\lr = 1, \dots, N$, $k=1, \dots, d$

\STATE 
Optimize scale parameters $\siglk$ of $f_1$ by minimizing $\hat R (f) $ subject to the constraints $\siglk=\sigk$, $ c^k = (\Phi^k)^{-1} y^k$

\STATE 
Estimate class labels $C_i$ and compute confidence scores $\mu_i$ for $i=1, \dots, Q$ by NN classification with $f_1$ in $\Rd$

\ENDFOR

\FOR {$\itr= 2, \dots, \Itr$}
\label{alg:mainloop_begin}

\STATE
Determine $\{ x_{\itr_\lr} \}_{\lr=1}^{\Lr_\itr}$ such that $\{ x_{\itr_\lr} \}_{\lr=1}^N= \X_T$ and $\{ x_{\itr_\lr} \}_{\lr=N+1}^{\Lr_\itr}$ are the points in $\X \setminus \X_T$ with highest confidence scores

\STATE
Set kernel centers as $\alk = x_{\itr_\lr}$ for $\lr = 1, \dots, \Lr_\itr$, $k=1, \dots, d$ 

\STATE
\label{alg:comp_proj}
Compute the embeddings of the projections of $ x_{\itr_\lr}$ on the manifolds as in \eqref{eq:embedding_Pmx} and set $y_{\itr_\lr} = \E \left( P_{\M_{C_{\itr_\lr}}}(x_{\itr_\lr}) \right)$

\STATE
\label{alg:opt_scale}
Optimize scale parameters $\siglk$ of $f_r$ by minimizing $\hat R (f) $ subject to the constraints $\siglk=\sigk$, $ c^k = (\Phi^k)^{-1} y^k$

\STATE 
\label{alg:upd_class}
Update class labels $C_i$ and confidence scores $\mu_i$ for $i=1, \dots, Q$ with NN classification with $f_r$ in $\Rd$

\ENDFOR 
\label{alg:mainloop_end}

\STATE
\textbf{Output}:\\
Out-of-sample interpolation function $f=f_\Itr: \Rn \rightarrow \Rd$ given by
$\fk(x) = \sum_{\lr=1}^{Q}  \cl^k \, \phi \left(  \frac{\| x -  \al^k \|}{\sigl^k} \right)$  \\
$\{ C_i \}_{i=N+1}^Q $: Class labels of initially unlabeled data samples
 
\end{footnotesize}

\end{algorithmic}
\label{alg:sosi}
\end{algorithm}

\section{Discussion}
\label{sec:discussion}

\subsection{Complexity analysis}

We now derive the complexity of the proposed method, which is essentially determined by the complexity of steps \ref{alg:comp_proj}-\ref{alg:upd_class} in the main loop of the algorithm. In step \ref{alg:comp_proj}, the determination of the nearest neighbors in $\X_T$ for each test image is of complexity $O(n N)$, and the solution of the quadratic program in \eqref{eq:quad_prog_embed} has a polynomial-time complexity $O(\poly(K))$ in the number of neighbors $K$ \cite{KozlovTK80}. The complexity $dK$ of \eqref{eq:embedding_Pmx} can be neglected as $d$ is small. Since the embedding of the projection of each point in $\X \setminus \X_T$ is computed only once throughout the algorithm, we get the overall complexity of step  \ref{alg:comp_proj} as $O(Q \, (\poly(K) + n N)) \approx O(n  Q N) $. 

Next, step \ref{alg:opt_scale} requires the evaluation of the regularization term $\hat R (f)$ at several $\sigk$ values and the corresponding coefficients $ c^k = (\Phi^k)^{-1} y^k$. The computation of the coefficients $c^k$ requires the solution of an $\Lr_\itr \times \Lr_\itr$ linear system, whose complexity is between $O({\Lr_\itr}^2)$ and  $O({\Lr_\itr}^3)$. Then, for a given $\sigk$ and the corresponding $c^k$, we analyze the evaluation of $\hat D (\fk)$. The computation of the gradient $\nabla \fk (x_i)$ is of complexity $O(n \Lr_\itr)$. Assuming that each training point $x_i$ has around $K$ nearest neighbors in each one of the $M$ classes, the computation of the directional derivative $\nabla_u \fk (x_i)$ for all neighbors of a point $x_i$ is of complexity $O( n ( \Lr_\itr + K M) )$. Since this is repeated for all $N$ training points $x_i$, the complexity of computing $\hat D (\fk)$ is of $O( n N ( \Lr_\itr + K M) )$. Since the complexity of $\hat G(\fk)$ is dominated by that of $\hat D (\fk)$, the optimization of $\sigk$ is of $O({\Lr_\itr}^2+  n N ( \Lr_\itr + K M) )$. Performing this optimization for all $d$ dimensions, upper bounding $\Lr_\itr$ by $Q$, and repeating this for all $\Itr$ iterations gives the complexity of step \ref{alg:opt_scale} throughout the algorithm as $O(  d \Itr ( {Q}^2+  n N ( Q+ K M) )) \approx O(  d n \Itr  N ( Q+ K M) )$. If one omits the reoptimization of $\sigk$ for $r>1$, the complexity of step \ref{alg:opt_scale} is reduced to the optimization of scale parameters at the first iteration $\itr=1$ and the update of the coefficients $c^k$ at every iteration, which is of $O(  d  n N ( N+ K M) + d \Itr {Q}^2)$.

Step \ref{alg:upd_class} requires the evaluation of $f(x_i)$ for all $x_i \in \X \setminus \X_T$, which is of $O( d n Q \Lr_\itr )$, and the comparison of the function values to those of the training points, which is of $O(d Q N)$. The complexity of repeating step \ref{alg:upd_class} throughout $\Itr$ iterations is then of $O( \Itr (d n  Q^2   + d Q  N) ) =  O( d n \Itr Q^2  ) $. Finally, combining the complexities of steps \ref{alg:comp_proj}-\ref{alg:upd_class}, we get the complexity of the overall algorithm as $O(d n \Itr Q^2 )$.

\subsection{Relation to kernel ridge regression}

In this section, we discuss how out-of-sample extensions of supervised manifold learning methods with RBF interpolation can be interpreted within the context of kernel ridge regression. Ridge regression is a well-known statistical method that learns a linear function to model the dependency between a set of input training points $\{ x_i  \}_{i=1}^N \subset \Rn$ and the associated outputs $\{ y_i \}_{i=1}^N \subset \R^d$. For each dimension $y_i^k$ of the outputs $y_i = [y^1_i \dots y^d_i]$, the algorithm looks for a linear model $f^k(x)=w^T x $ that minimizes  
\begin{equation*}
G(w) = a \| w \|^2 + \sum_{i=1}^N (y^k_i - w^T x_i)^2
\end{equation*}
which is a slightly modified version of the least squares method by adding a regularization term representing the squared norm of the fitted linear model. Here $a>0$ is a parameter adjusting the weight of the regularization term. An alternative formulation of ridge regression is proposed in \cite{SaundersGV98} that is based on a dual version of the above problem. The solution of the dual problem yields the following prediction $\fk(x)$ of the output value for a new input sample $x$:
\begin{equation}
\label{eq:ridge_reg}
\fk(x)=(y^k)^T (K + a I )^{-1} v.
\end{equation}
Here, $y^k=[y^k_1 \dots y^k_N]^T$ is the vector of output values for training samples, $K \in \R^{N \times N}$ is the matrix of inner products of input samples whose entries are given by $K_{ij}= \langle x_i, x_j \rangle$, $I$ is the identity matrix, and $v \in \R^{N \times 1}$ is the vector of inner products of $x$ with $x_i$, whose $i$th entry is given by $v_i = \langle x, x_i  \rangle$. 

Since this formulation only involves the inner products between the samples $x$ and $\{ x_i \}$ rather than the samples themselves as vectors, it permits a kernel extension of the regression problem, where the samples are mapped to a high-dimensional feature space $F$ via a kernel $\psi: \Rn \rightarrow F$. The inner products in $K$ and $v$ are then evaluated in the feature space as $K_{ij} = \langle  \psi(x_i), \psi(x_j) \rangle$ and $v_i = \langle \psi(x) , \psi(x_i)   \rangle$. Translation-invariant kernels are a widely-used family of kernel functions, where the inner product $\langle  \psi(x_i), \psi(x_j) \rangle$ in the feature space depends only on the difference $\| x_i - x_j \|$ between the samples in the original space. 

Out-of-sample extensions with RBF kernels as in the proposed method are linked to kernel ridge regression in the following way. If the regularization term ($a=0$) is omitted in \eqref{eq:ridge_reg}, the $k$th dimension of the output vector for the input sample $x$ is given by
\begin{equation}
\label{eq:ridgereg_noregul}
\fk(x)=(y^k)^T K ^{-1} v.
\end{equation}
If the kernel $K_{ij}$ is set as
$
 \langle  \psi(x_i), \psi(x_j) \rangle = \phi (   \|  x_i  - x_j \| / \sigma  )
$
with the RBF kernel used in interpolation, one can observe from \eqref{eq:defn_Phi_matrix} that the kernel matrix $K$ coincides with the matrix $\Phi^k$ when a constant scale parameter $\sigma$ is chosen for dimension $k$ of the interpolation function. Defining $v$ similarly with the RBF kernel $\phi$, the interpolation function in \eqref{eq:int_func_form} can be written as $f^k (x) = (c^k)^T v$. The coefficients $c^k$ of the interpolation function being given by $c^k = (\Phi^k)^{-1} y^k$, we obtain
\begin{equation*}
f^k (x) = (c^k)^T v =   (y^k)^T (\Phi^k)^{-1} v 
\end{equation*}
which is the same as the result obtained with kernel ridge regression in \eqref{eq:ridgereg_noregul}. 

We thus observe that fitting an RBF interpolation function for manifold embeddings is the equivalent of learning a kernel ridge regression model (with no regularization) such that the output values $y^k_i$ are the coordinates of data samples in the computed embedding. Therefore, the studied out-of-sample extension setting can be regarded as a kernel ridge regression adapted particularly to manifold-structured data. Indeed, in the general and traditional regression setting for classification, no assumption is made about the structure of data, and the output vectors $y_i$ are taken as the class labels.  Taking $y_i$'s simply as the class labels of data transmits only the class information to the regression algorithm and conveys no information about the geometric properties of data. Meanwhile, first computing an embedding with a supervised manifold learning algorithm and then learning the regression model on the coordinates $y_i^k$ of data in $\Rd$ (instead of taking $y_i$'s directly as class labels)  allows the classifier to be guided by the special geometric structure of data samples concentrated around class-representative manifolds. Coordinates learned with supervised manifold learning algorithms reinforce the class information of data by enhancing the separability between the classes, while the manifold structure of data is also preserved in each class.



\section{Experimental results}
\label{sec:exp_results}

In this section, we evaluate the performance of the proposed method in classification experiments. We apply the presented out-of-sample extension algorithm on two different supervised manifold learning methods. First, we consider the supervised Laplacian eigenmaps algorithm presented in \cite{Raducanu12}, which computes an embedding by solving \eqref{eq:obj_supLapEmb}. Next, we evaluate our algorithm on embeddings obtained with the Fisher-like objective function in \eqref{eq:obj_supFisherEmb}, which is used by methods such as \cite{Hua12}, \cite{Yang11}, \cite{Zhang12}, and \cite{Wang09}. However, we compute a nonlinear embedding by removing the linear projection constraint $z^T=v^T X$, so that the out-of-sample extension problem is of interest.  

We compare the following methods in the experiments, the first four of which provide out-of-sample extension solutions for manifold embeddings. When testing the out-of-sample extension methods, class labels of test images are assigned with nearest-neighbor classification in the low-dimensional domain of embedding. 
\begin{itemize}
\item Proposed semi-supervised out-of-sample interpolation method (SOSI)
\item RBF fitting: An RBF interpolation function is fitted only to the training samples, which is the equivalent of the interpolation function $f_1$ computed at the end of iteration $r=1$ in Algorithm \ref{alg:sosi}. Test images $x$ are then mapped to $\R^d$ via the function $f_1(x)$.
\item Locally linear embedding (LLE): Test points in $\Rn$ are mapped to $\Rd$ with an adaptation of the LLE algorithm \cite{Roweis00} to the out-of-sample problem. Given a test point $x \in \Rn$, first its approximation is computed as a linear combination of its nearest neighbors in $\X_T$ with weights adding up to $1$ as in LLE. The point $x$ is then mapped to $y \in \Rd$ as the linear combination of the embeddings of the same neighbors with the same weights.
\item Nystr\"om: The original Nystr\"om formula is not applicable since the data-dependent kernel depends on the class labels as discussed in Section \ref{ssec:overv_oos_ext}. We thus use a modified version of the Nystr\"om method, where $\fk(x)$ is taken as a linear combination of the embedding coordinates $y_i^k$ weighted by the kernel as in \eqref{eq:Nystrom}. The kernel $\tilde M$ in the formula is taken as the same type of kernel (Gaussian kernel) used in the construction of the within-class and between-class weight matrices $W_w$ and $W_b$, and it is normalized for each test sample so that the kernel values $\tilde M(x, x_i)$ sum up to $1$.
\item Nearest neighbor classification in the original data space $\Rn$
\item SVM in the original data space $\Rn$
\item Semi-supervised learning (SSL) using Gaussian fields: Since the proposed out-of-sample extension method can be regarded as a building block of a semi-supervised classifier, we also compare our results with those of a semi-supervised classification method. We test the performance of SSL with the algorithm proposed in \cite{ZhuGL03}, which is a state-of-the-art semi-supervised classifier based on the computation of a smooth function on the data graph that coincides with the class labels when evaluated at data samples of known class labels.

\end{itemize}

\begin{figure}[t]
\begin{center}
     \subfigure[Supervised Laplacian eigenmaps]
       {\label{fig:embedLap}\includegraphics[height=4cm]{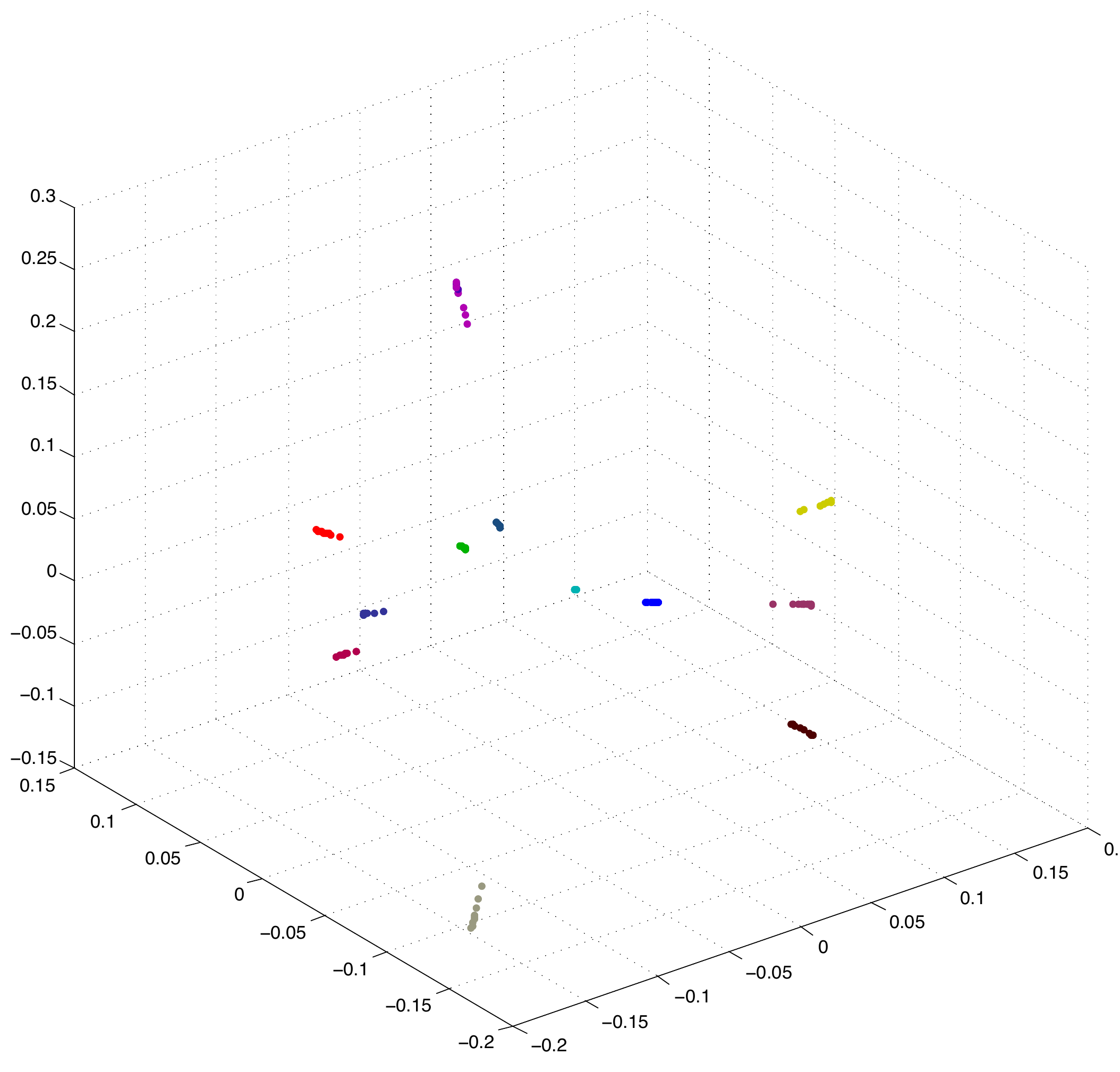}}
     \subfigure[Fisher-based embedding]
       {\label{fig:embedFish}\includegraphics[height=4cm]{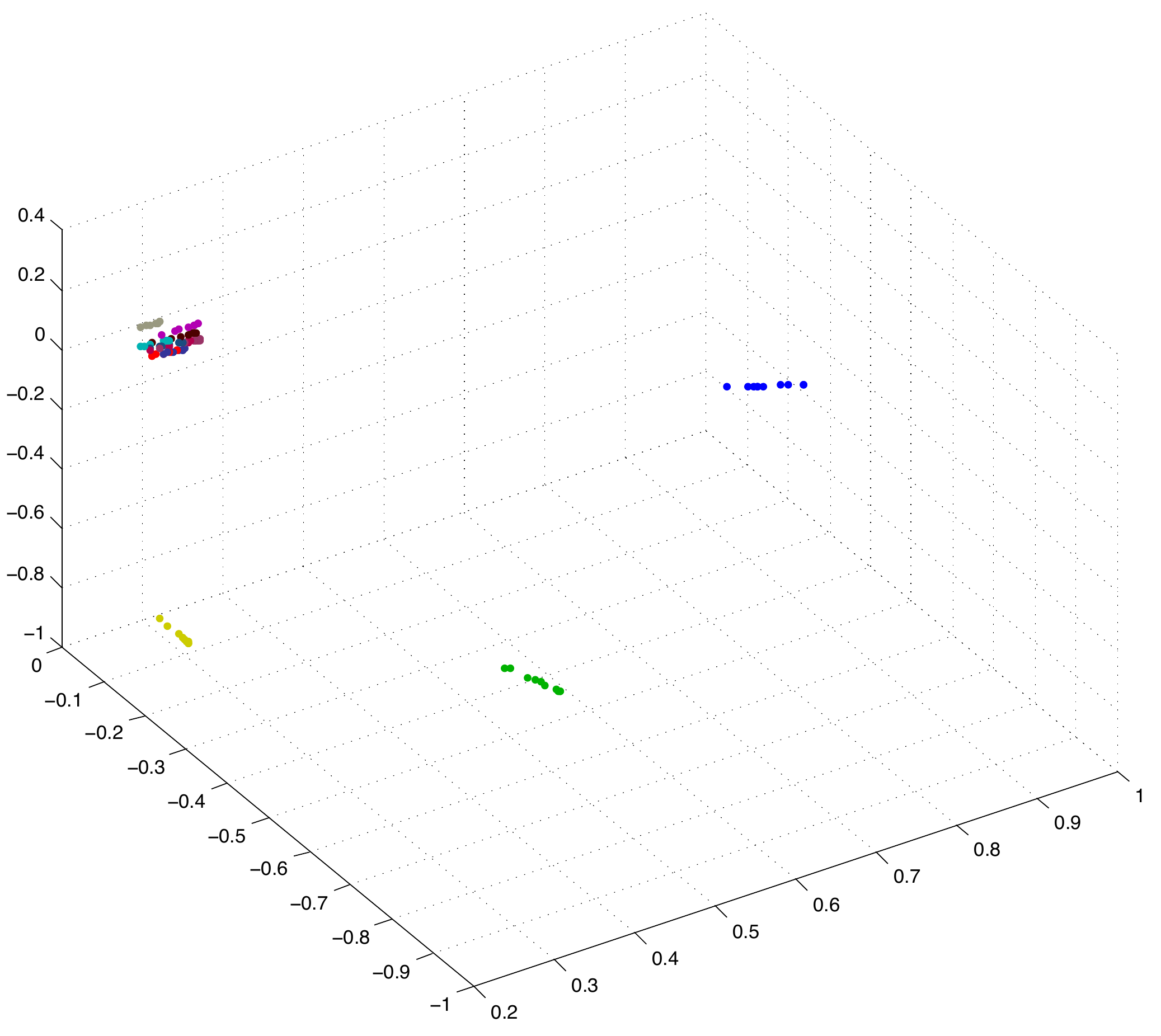}}
 \end{center}
 \caption{Three-dimensional embeddings of the Yale face data set obtained with the two manifold learning methods used in the experiments}
 \label{fig:embeddingsYale}
 \vspace{-17pt}
\end{figure}

\begin{figure}[t]
\begin{center}
     \subfigure[Yale face database]
       {\label{fig:datasetYale}\includegraphics[height=2.5cm]{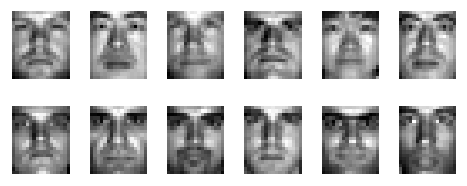}}
     \subfigure[ETH-80 object database]
       {\label{fig:datasetETH}\includegraphics[height=1cm]{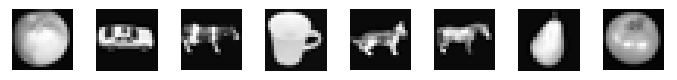}}
     \subfigure[COIL-20 object database]
       {\label{fig:datasetCoil}\includegraphics[height=1.8cm]{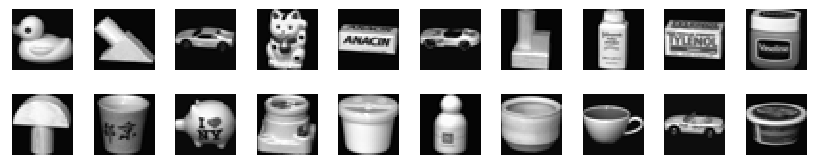}}
 \end{center}
 \caption{Sample images from data sets used in the experiments}
 \label{fig:datasets}
 \vspace{-17pt}
\end{figure}

\begin{figure}[t]
\begin{center}
     \subfigure[Supervised Laplacian eigenmaps]
       {\label{fig:errorsYaleLap}\includegraphics[height=3.3cm]{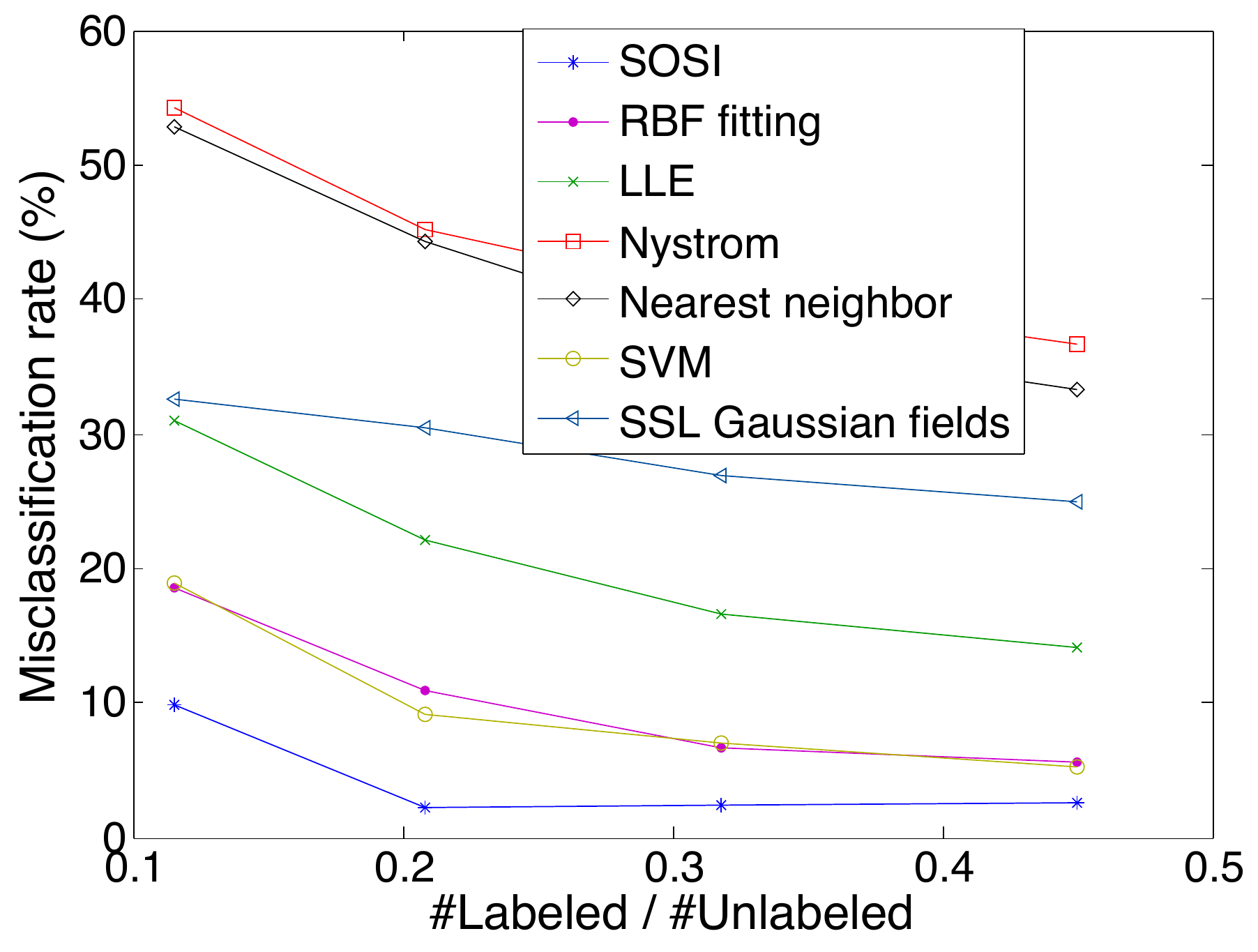}}
     \subfigure[Fisher-based embedding]
       {\label{fig:errorsYaleFish}\includegraphics[height=3.3cm]{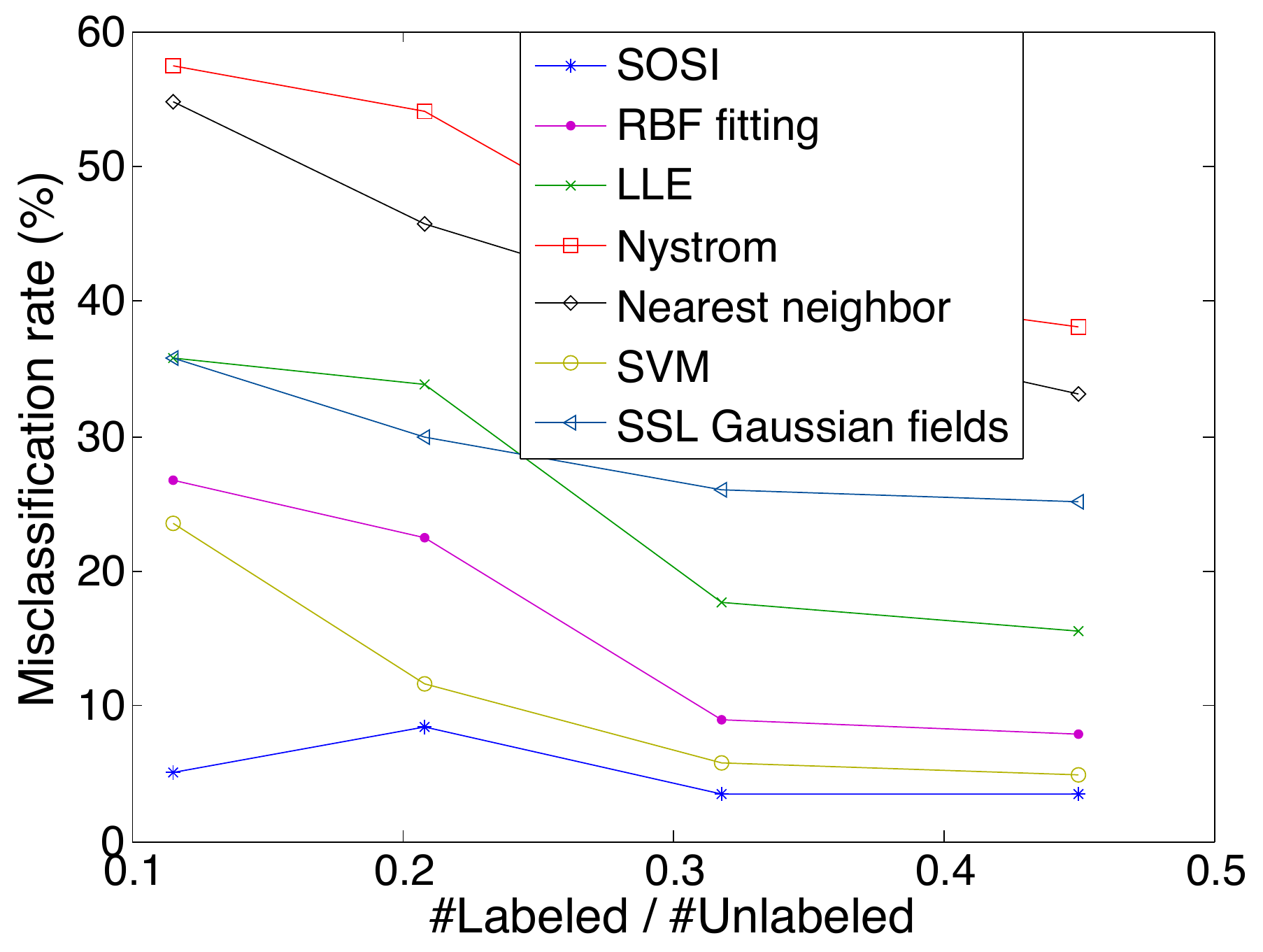}}
 \end{center}
 \caption{Misclassification rates of face images from Yale database}
 \label{fig:errorsYale}
\end{figure}

We first evaluate the proposed method on a data set consisting of the face images of 12 individuals from the extended Yale face database \cite{GeBeKr01}, which includes 58 images of each individual taken under different poses and illumination conditions. The images are normalized, converted to grayscale and downsampled to a resolution of $17 \times 20$ pixels. A sample image of each subject in the data set is shown in Figure \ref{fig:datasetYale}. The supervised Laplacian eigenmaps and the Fisher-based embedding algorithms are used to map the data ($17 \times 20$-pixel images) to $\R^{20}$. The weight parameter is set as $\mu=0.01$ in the supervised Laplacian eigenmaps method. Figure \ref{fig:embeddingsYale} shows the embeddings of a subset of the data set containing 10 labeled images of each individual, computed with the supervised Laplacian and the Fisher-based embedding algorithms. Only the first three dimensions of the coordinates are plotted for illustration. It can be observed that both methods compute representations with an enhanced separation between different classes. The supervised Laplacian eigenmaps method yields an even distribution of different classes across different dimensions. Since each dimension of the embedding renders several pairs of classes separable, sufficiently many class pairs contribute to the total directional derivative $\hat D(\fk)$ in \eqref{eq:regobj_disc_D} for each dimension $k$. This causes the variations of $\hat D(\fk)$ and $\hat G(\fk)$ with the scale parameter to be as discussed in Section \ref{ssec:interp_comp}, such that $\hat G(\fk)$ increases at a faster rate than $\hat D(\fk)$ at large scales due to overfitting. Thus, for the embeddings obtained with supervised Laplacian eigenmaps, we optimize the scale parameters by minimizing the regularization term $\hat R(f)$ as in \eqref{eq:reg_obj_disc}.\footnote{Occasionally, the scale parameter $\sigk$ of one dimension or a few dimensions $k$ may diverge from the scale parameters of the rest of the dimensions, which may cause instabilities. In order to avoid this, we bound the final values of the scale parameters to an interval of two standard deviations around their mean value averaged over all dimensions.} Meanwhile, the embedding computed with the Fisher-based objective yields a more ``polarized'' representation, where each dimension of the embedding is observed to separate out only one class from the others. When there are not sufficiently many separable class pairs in $\hat D(\fk)$, the estimation of the variation of this term with the scale parameter may become unreliable or biased by a particular class in each dimension. We have observed that, when the embedding is computed with the Fisher-based objective, the variation of $\hat D(\fk)$ with the scale parameter is closer to that of $\hat G(\fk)$ (in comparison with supervised Laplacian eigenmaps). 
The choice of the regularization term $\hat R (f)$ as a linear combination of these two terms may then lose its reliability, as it may become a monotonic function of the scale parameter, for instance. Therefore, for the Fisher-based embedding, we apply a slightly modified procedure for optimizing the scale parameters, where we choose a sufficiently large value for the scale parameter in each dimension, which ensures, however, that the $\hat D(\fk) / \hat G(\fk)$ ratio stays above a certain threshold value. The scale parameters of the RBF fitting method are set as equal to those of the proposed SOSI algorithm. Figure \ref{fig:errorsYale} shows the classification errors obtained with all methods for the supervised Laplacian and the Fisher-based embeddings. Each curve displays the misclassification rate (in percentage) of unlabeled images, obtained by varying the ratio between the number of labeled and unlabeled images in the data set. The results are the average of 5 repetitions of the experiment by randomly choosing the labeled samples. An early stopping rule is applied in the SOSI algorithm for the leftmost point of the curve (the labeled/unlabeled ratio of 0.11) due to the relatively high error, where the interpolation function construction is terminated when around $80\%$ of the unlabeled points are added as RBF kernel centers. It is observed that the proposed method outperforms the other out-of-sample extension methods in comparison, as well as the SVM classifier and the semi-supervised graph-based classifier.

\begin{figure}[t]
\begin{center}
     \subfigure[Supervised Laplacian eigenmaps]
       {\label{fig:errorsETHLap}\includegraphics[height=3.25cm]{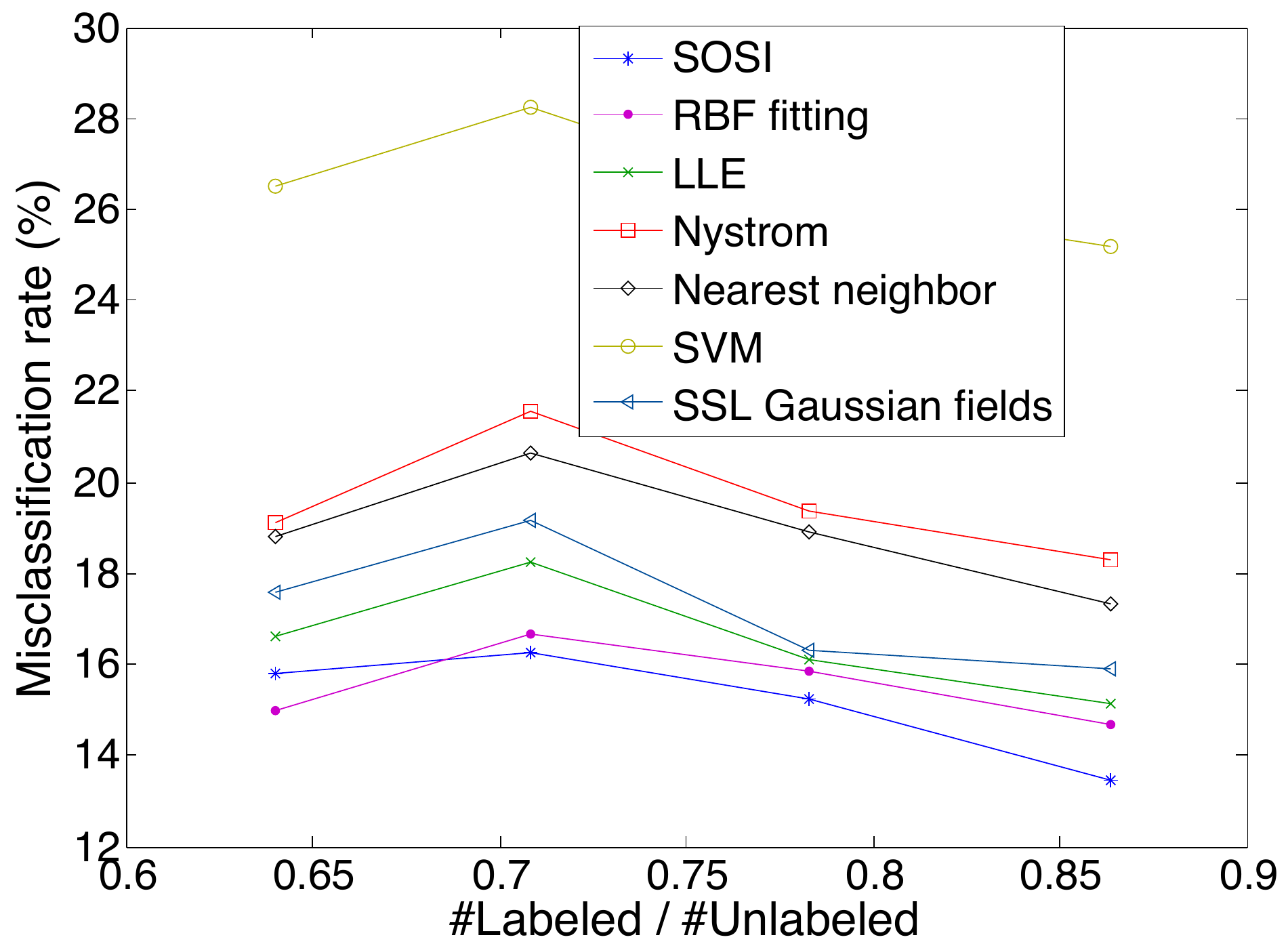}}
     \subfigure[Fisher-based embedding]
       {\label{fig:errorsETHFish}\includegraphics[height=3.25cm]{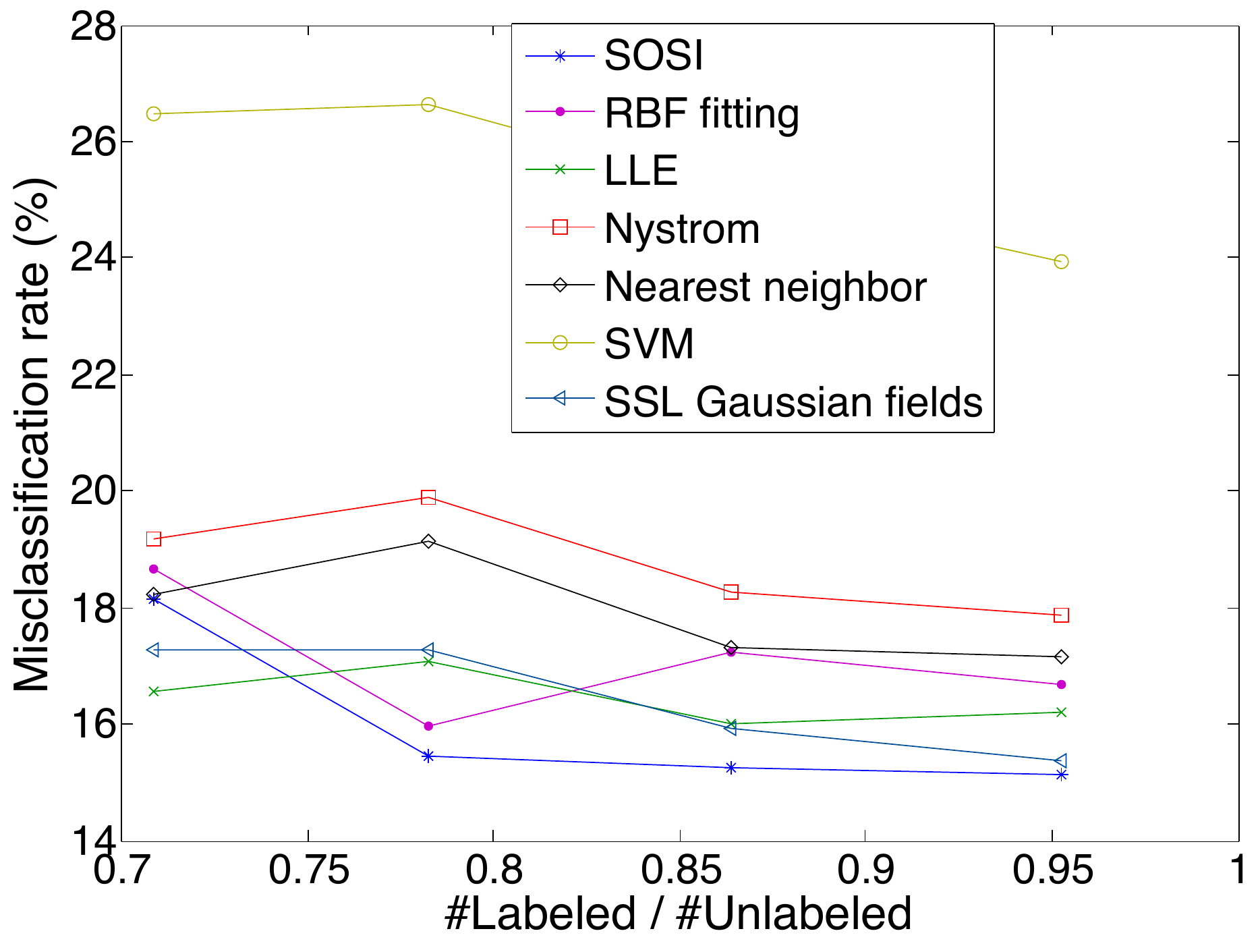}}
 \end{center}
  \caption{Misclassification rates of object images from ETH-80 database}
  \label{fig:errorsETH80}
\end{figure}
\begin{figure}[t]
\begin{center}
     \subfigure[Supervised Laplacian eigenmaps]
       {\label{fig:errorsCoilLap}\includegraphics[height=3.8cm]{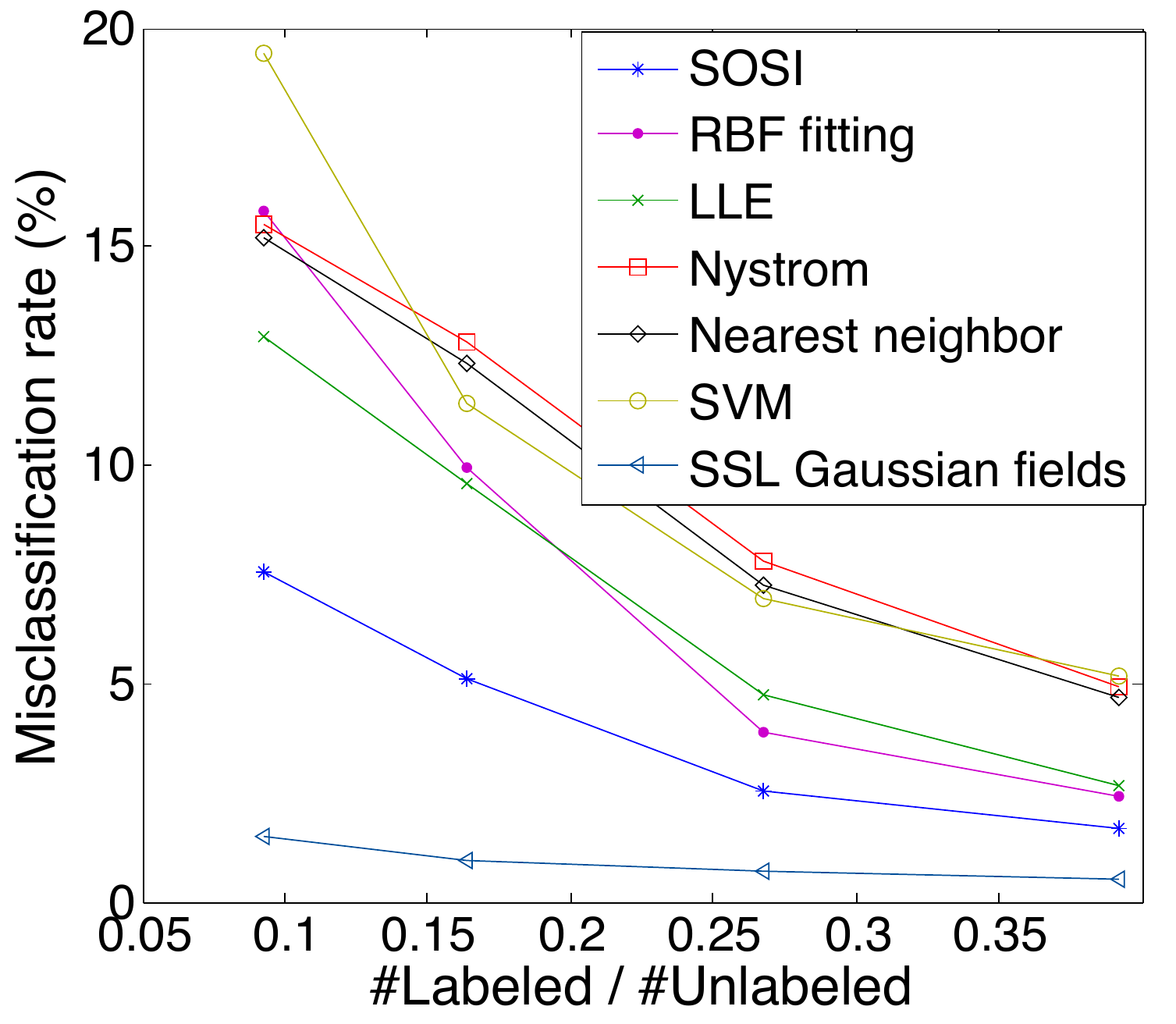}}
     \subfigure[Fisher-based embedding]
       {\label{fig:errorsCoilFish}\includegraphics[height=3.8cm]{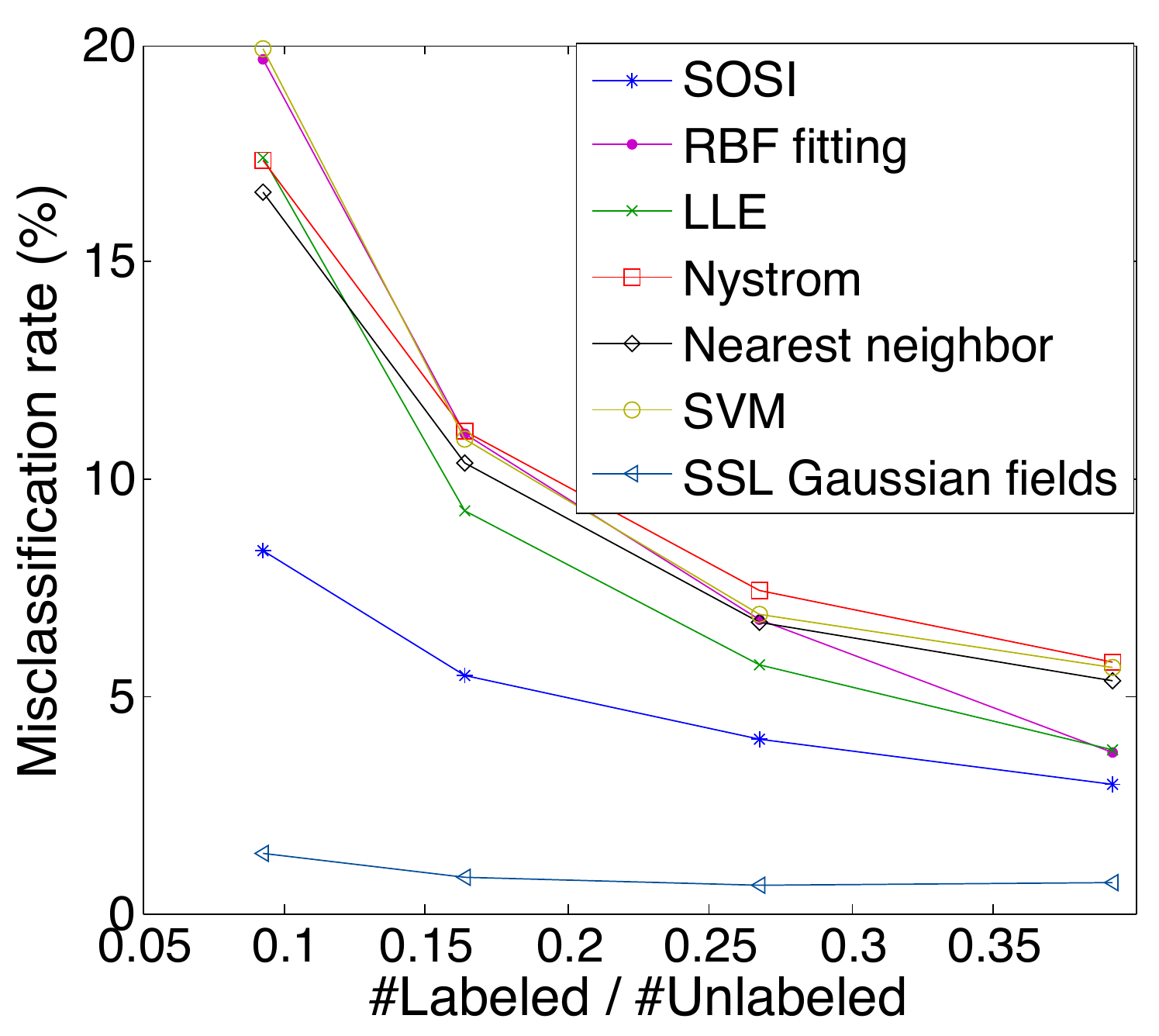}}
 \end{center}
  \caption{Misclassification rates of object images from COIL-20 database}
  \label{fig:errorsCOIL20}
\vspace{-15pt}
\end{figure}


We then repeat the same experiment on two different databases of object images captured under varying viewpoints. The first experiment is conducted on the images of 8 objects from the ETH-80 database \cite{LeibeS03}, where $41$ images are available for each object (in particular, the images of the first object in each object category are used so that the images in each class belong to the same manifold). A sample image of each object is shown in Figure \ref{fig:datasetETH}. The images are normalized, converted to grayscale, and downsampled to a resolution of $20 \times 20$ pixels. An embedding of dimension $d=15$ is computed with the supervised Laplacian eigenmaps and the Fisher-based manifold learning algorithms.  The second experiment is done on the images of 20 objects from the COIL-20 database \cite{NeneNM96} with $71$ images for each object, which are normalized, converted to grayscale, and downsampled to a resolution of $32 \times 32$ pixels. Figure \ref{fig:datasetCoil} shows a sample image for each object. The images are embedded in a space of dimension $d=25$. In both experiments, the optimization of the scale parameters is done as in the previous experiment. The results obtained with the two object data sets are presented in Figures \ref{fig:errorsETH80} and \ref{fig:errorsCOIL20}. The misclassification rates of unlabeled samples are plotted with respect to the ratio between the number of labeled and unlabeled samples. The results are the average of $5$ random partitionings of the data set. As the classification error of the ETH-80 database is relatively high, an early stopping rule is applied for this data set by terminating the interpolation function construction when around $70 \%$ of the unlabeled samples with the highest confidence scores are added as RBF kernel centers. The results show that the proposed method often yields the smallest classification error in the experiment of Figure \ref{fig:errorsETH80}, while it is outperformed only by the semi-supervised learning method in Figure \ref{fig:errorsCOIL20}. This graph-based semi-supervised learning algorithm performs particularly well on the COIL-20 data set, due to the dense sampling and the regular structure of the object image manifolds.

The overall consideration of these experiments shows that the proposed out-of-sample extension method for supervised manifold learning provides a better performance than the reference out-of-sample extension strategies in comparison, while it can provide an alternative solution for semi-supervised learning when coupled with a supervised dimensionality reduction method and thus regarded as one building block of a semi-supervised classifier. In particular, one can observe in Figures \ref{fig:errorsYale}-\ref{fig:errorsCOIL20} that SVM and graph-based SSL may perform very differently in different settings. SVM is based purely on the representation of the data samples in the original ambient space $\Rn$, while graph-based SSL only uses the information of the similarities between neighboring data samples instead of interpreting them as vectors in the high-dimensional space $\Rn$. Meanwhile, the proposed method is expected to find a compromise between these two approaches, as the interpolation function depends both on the coordinates of the data samples in $\Rn$ and the coordinates of the embedding in $\Rd$ learned with a supervised manifold learning algorithm that relies on the graph representation of data. The experimental results seem to confirm this expectation, as the proposed classification solution attains a reasonably good performance in situations where SVM or graph-based SSL may fail (as in Figures \ref{fig:errorsETH80} and \ref{fig:errorsYale} respectively).

\begin{figure}[t]
\begin{center}
     \subfigure[Supervised Laplacian eigenmaps]
       {\label{fig:evolYaleLap}\includegraphics[height=3.7cm]{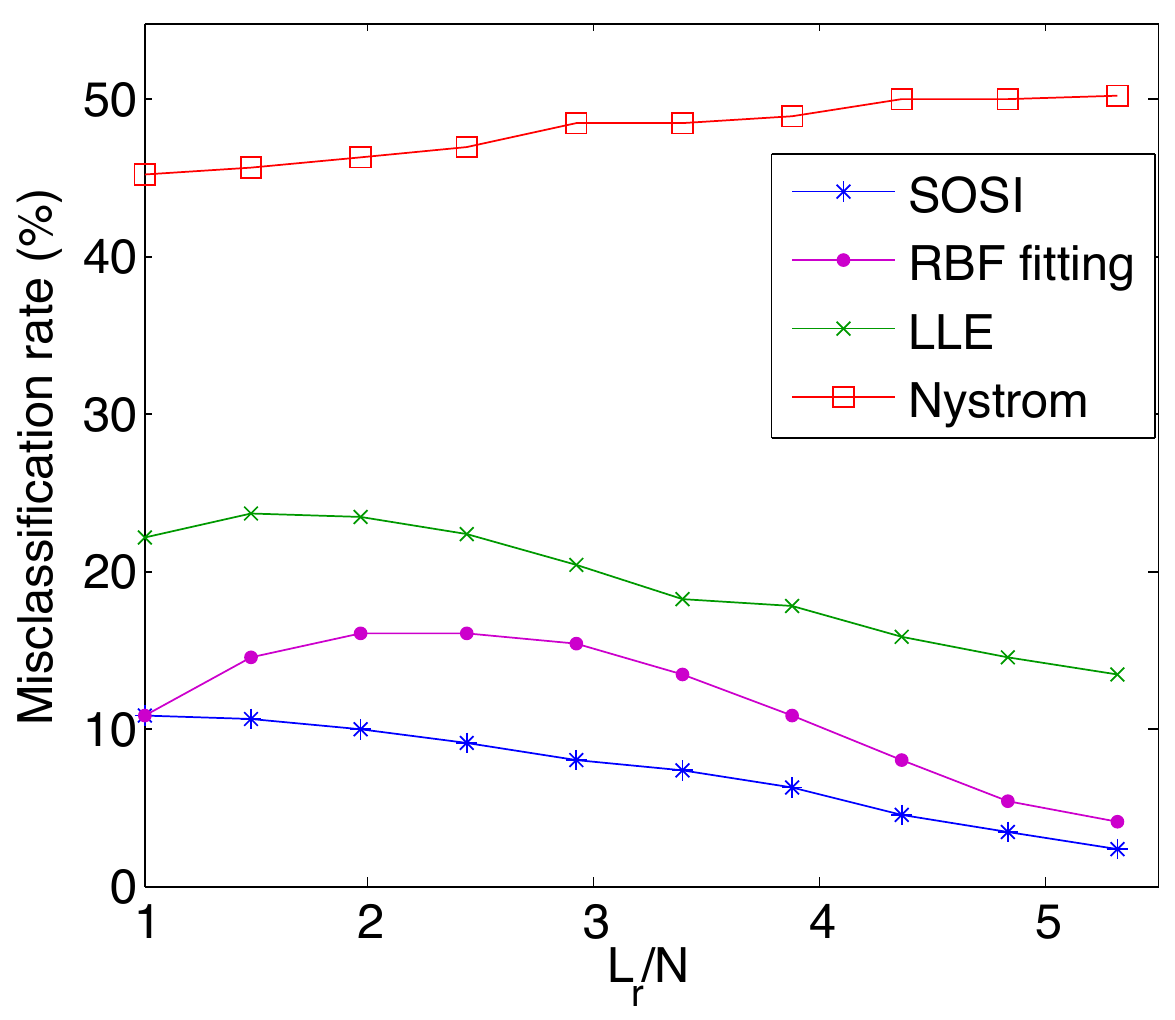}}
     \subfigure[Fisher-based embedding]
       {\label{fig:evolYaleFish}\includegraphics[height=3.7cm]{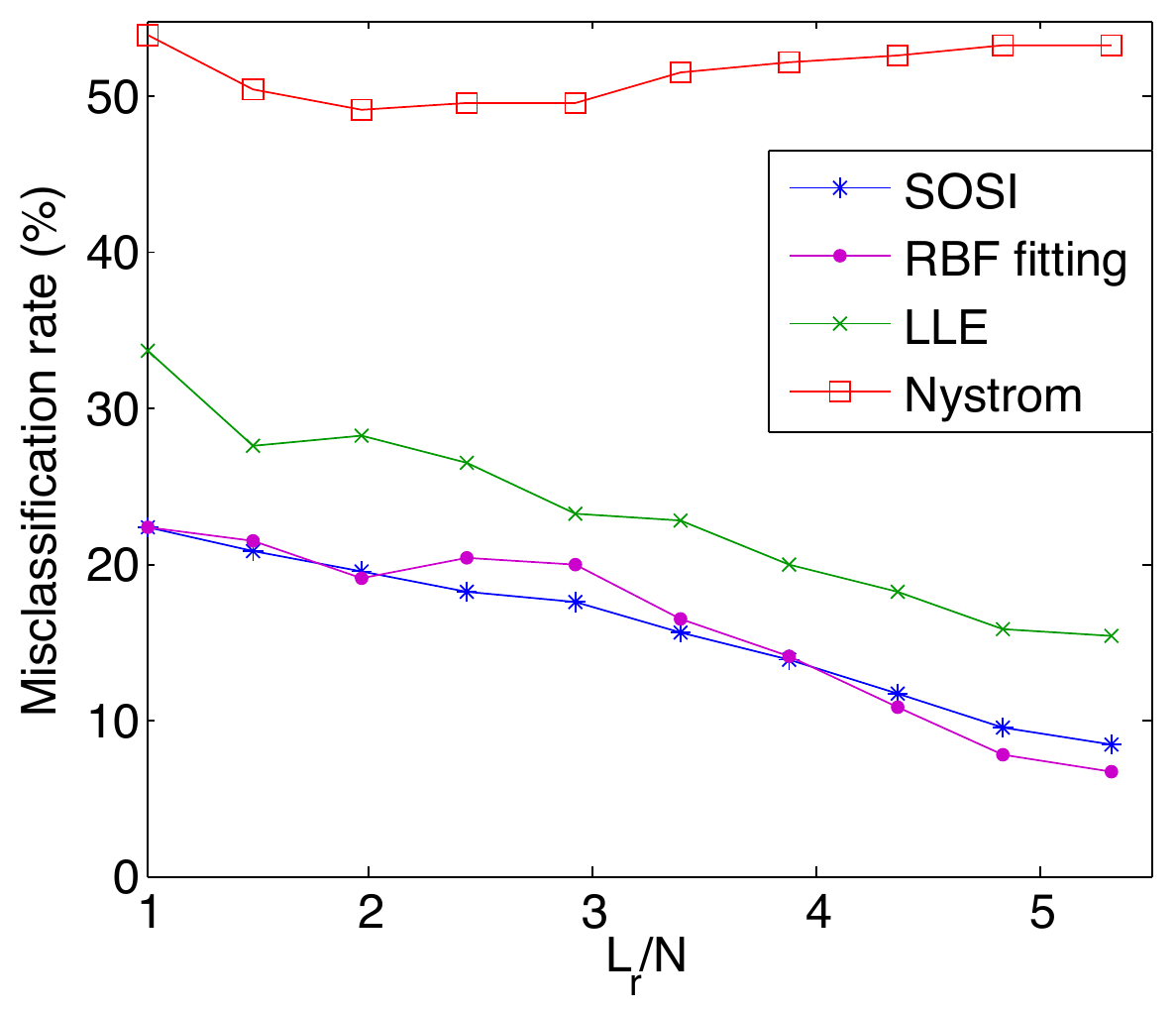}}
 \end{center}
 \caption{Misclassification rates obtained with progressive integration of the test images in the extended training set. Out-of-sample extensions are computed, class labels are assigned, and embeddings are updated with the extended training set in each iteration.}
 \label{fig:evolutionYale}
\vspace{-20pt}
\end{figure}

In the experiments of Figures \ref{fig:errorsYale}-\ref{fig:errorsCOIL20}, the interpolation functions of the out-of-sample methods other than SOSI are constructed using only the training data. The information present in the unlabeled data samples is not exploited in the construction of these interpolation functions, whereas SOSI uses these points to gradually add them as kernel centers of the learned interpolation function. In order to assess the performance of the proposed method in the progressive integration of the unlabeled data samples in the learning process, we do an additional experiment. The proposed SOSI algorithm is used to classify unlabeled test images in an iterative way as described in Algorithm \ref{alg:sosi}. Then, in order to compare SOSI with the other out-of-sample extension methods, for each one of these methods, we carry out an iterative classification procedure as follows. In each iteration, all test images are assigned class labels with nearest-neighbor classification in the low-dimensional domain via the out-of-sample generalization strategies in comparison, and a confidence score is obtained for each test image as in \eqref{eq:confid_meas_defn}. Then in the next iteration, the test images with the highest confidence scores are added to the training set with their estimated class labels and a completely new embedding of this  extended training set is computed with the supervised Laplacian eigenmaps and the Fisher-based embedding algorithms (thus new coordinates are assigned to the original training images as well). The out-of-sample extension of this new embedding is then recomputed with the tested strategies in comparison, which are used to reclassify the test images. In each iteration $\itr$, the compared methods use the same number $\Lr_\itr$ of extended training images in $\X$ (same as the number of terms in the interpolation function of SOSI), while the choice of the extended training set varies between the compared methods as a result of the different confidence scores they assign to the test images. This progressive procedure is continued until all test images are included in the extended training set. The results obtained on the face images from the Yale database are presented in Figures \ref{fig:evolYaleLap} and \ref{fig:evolYaleFish}, respectively for the supervised Laplacian eigenmaps and the Fisher-based embedding algorithms. The image set of each subject contains 10 labeled and 48 unlabeled samples in this experiment. The misclassification rates obtained throughout the iterations are plotted with respect to the ratio $\Lr_\itr / N$ between the size of the extended training set (number of RBF terms for SOSI) and the size of the original training set. The results indicate that the best classification accuracy is achieved by the proposed algorithm most of the time. The misclassification error obtained with the proposed method decreases regularly throughout the iterations as the number of terms in the interpolation function increases, while the evolution of the misclassification error with the other strategies is less regular and the error may even increase throughout the iterations. This is due to the fact that the strategies other than SOSI compute a new embedding of the extended training set from scratch in each iteration. The mislabeled data samples in the extended training set may then significantly influence the computed embedding and consequently the class label assignments of the next iteration, since the embedding given by the eigenvectors of a class-dependent kernel matrix may change dramatically even with small errors in the kernel matrix. The proposed method does not suffer from this problem, since it preserves the original embedding and refines only the interpolation function throughout the iterations, which has a regularizing effect that better tolerates inaccurate assignments of the class labels of test images. We finally note that, among the strategies compared in this experiment, the proposed SOSI algorithm is the only one that provides an out-of-sample solution for manifold learning when $\Lr_\itr > N$.

\begin{figure*}[t]
\begin{center}
     \subfigure[Yale face data set]
       {\label{fig:reg_class_Yale}\includegraphics[height=3.5cm]{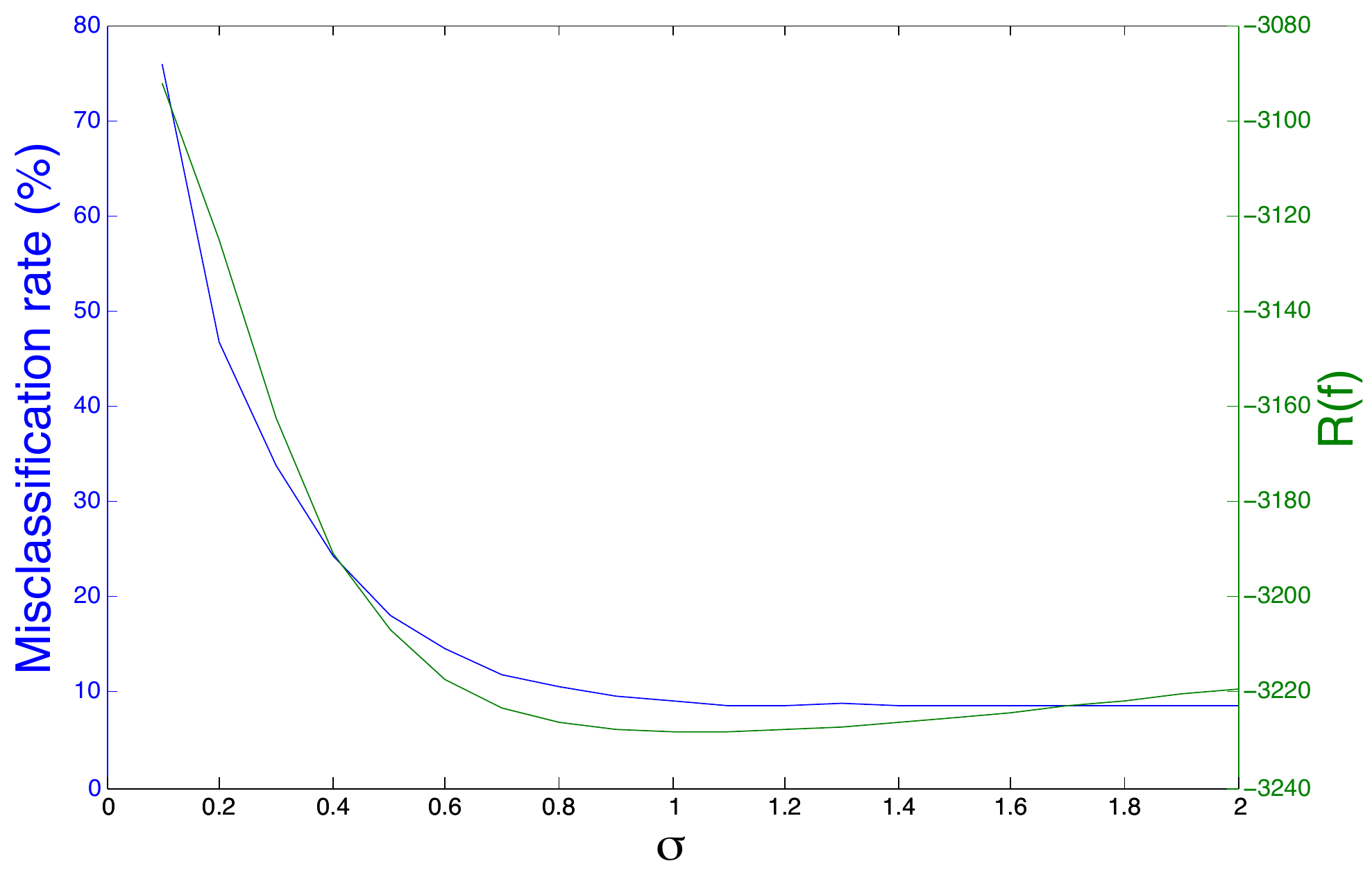}}
     \subfigure[ETH-80 object data set]
       {\label{fig:reg_class_ETH}\includegraphics[height=3.5cm]{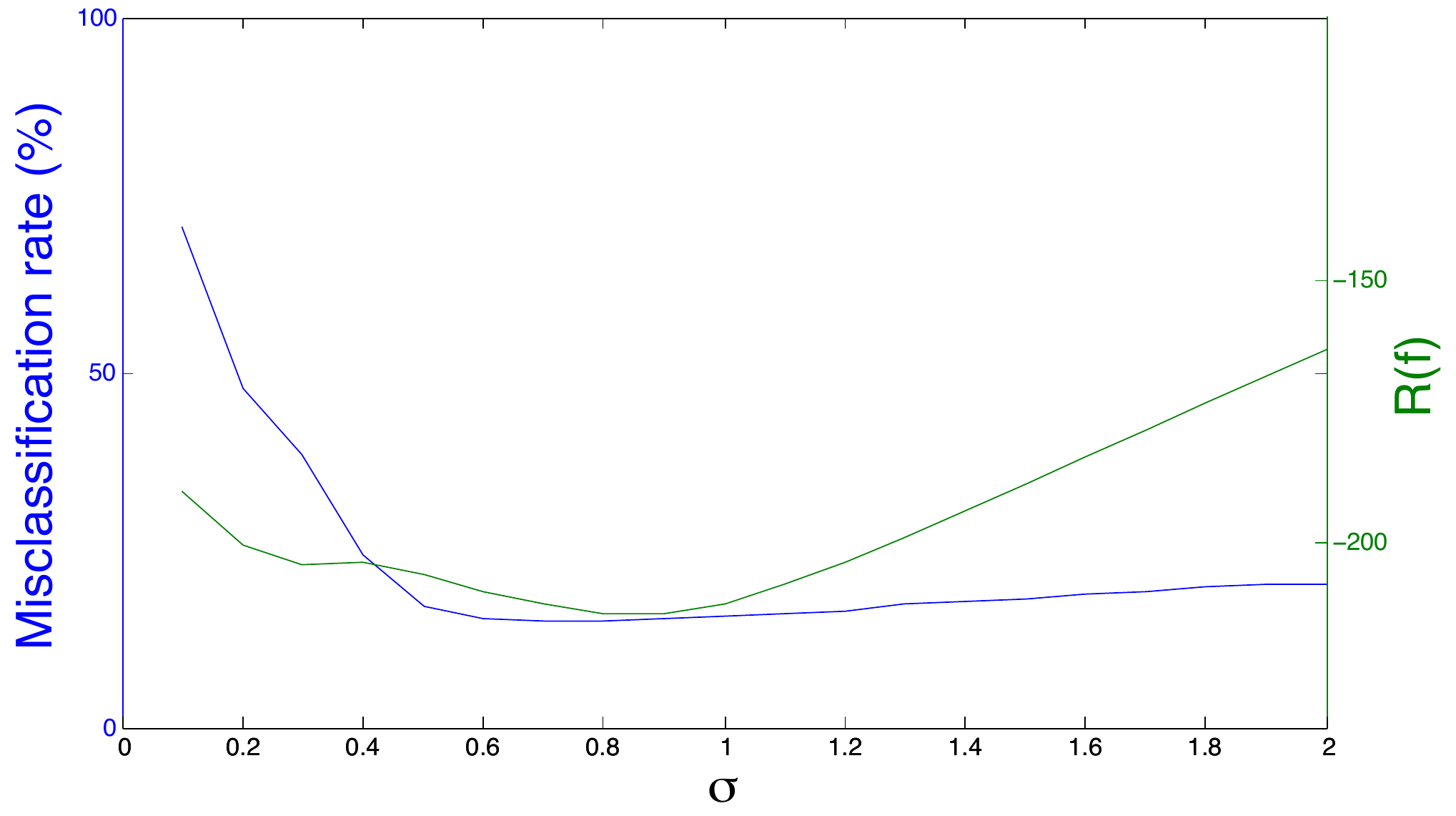}}
     \subfigure[COIL-20 object data set]
       {\label{fig:reg_class_Coil}\includegraphics[height=3.5cm]{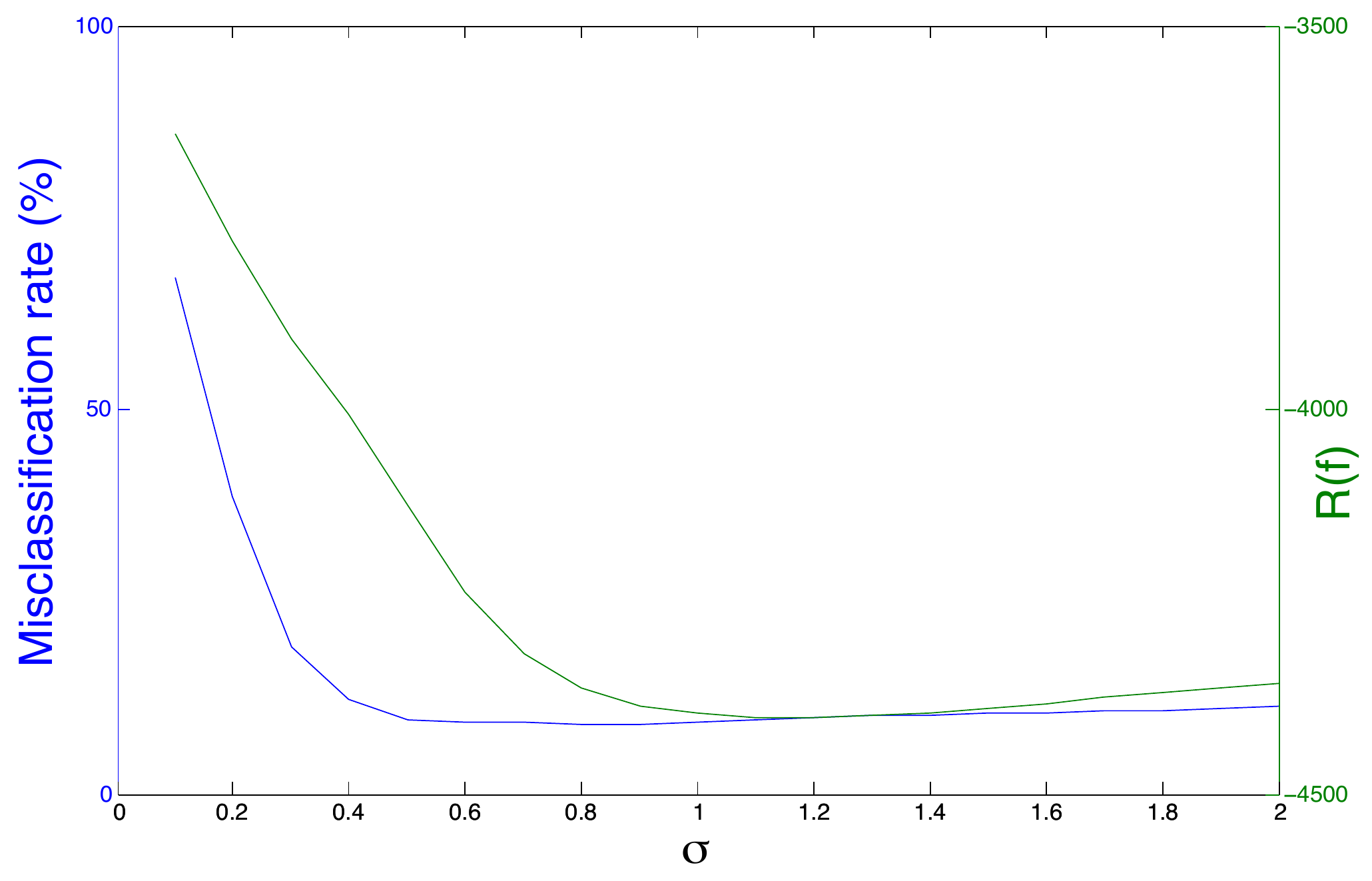}}
 \end{center}
 \caption{Variations of the misclassification error and the regularization term $\hat R(f) $ with the scale parameter of the RBF kernels}
 \label{fig:reg_class_variation}
 \vspace{-18pt}
\end{figure*}

Finally, we study the influence of the scale parameters of the interpolation function on the classification performance. As discussed in Section \ref{ssec:interp_comp}, the proposed method selects the scale parameters by optimizing the regularization term $\hat R (f)$. In order to evaluate the effect of this regularization approach on the classification accuracy, we compare the variations of the regularization term $\hat R (f)$ and the classification error with the scale parameter. We compute an embedding of the training images with the supervised Laplacian eigenmaps algorithm and then construct an RBF interpolation function, where all scale parameters $\siglk$ are set to a common $\sigma$ value and the coefficients $\clk$ are computed to fit  the training images and the learned embedding for this choice of the scale parameter (as in the RBF fitting method or the first iteration of SOSI). A sequence of interpolation functions are computed by varying the scale parameter $\sigma$, and for each interpolation function, the regularization objective $\hat R(f)$ is computed as well as the misclassification rate of the test images. The variations of the regularization cost and the misclassification rate with the scale parameter $\sigma$ are presented in Figure \ref{fig:reg_class_variation} for all three data sets used in the experiments. The results suggest that the regularization objective $\hat R(f)$ has a rather smooth and nonmonotonic variation with the scale parameter, which resembles that of the classification error. Moreover, the interval of scale parameters $\sigma$ minimizing the regularization objective $\hat R(f) $ coincides with the range of $\sigma$ values where the misclassification rate takes small values. This shows that the proposed regularization objective permits the algorithm to capture the influence of the scale parameters on the performance of learning and can be used for optimizing the scale parameters.

\section{Conclusions}
\label{sec:concl}

We have proposed a method for the out-of-sample extensions of supervised manifold learning algorithms that embed a set of class-representative manifolds residing in a high-dimensional ambient space to a set of manifolds in a lower-dimensional domain. The proposed out-of-sample generalization method is based on the construction of an RBF interpolation function, where the parameters of the interpolation function are optimized to minimize the embedding error over a set of initially unlabeled data samples, whose class labels are estimated progressively along with the parameters of the interpolation function. We have shown that the regularity of the interpolation function can be controlled by optimizing the RBF scale parameters to minimize a regularization objective that controls the total gradient of the interpolation function while encouraging sufficiently strong derivatives along the directions of class separation boundaries in order to ensure an effective separation between different classes. The proposed out-of-sample generalization method outperforms baseline interpolation solutions in classification applications. Experimental results suggest that the proposed algorithm achieves state-of-the-art performance in semi-supervised learning and can be effectively used along with supervised manifold learning methods in the classification of low-dimensional data sets consisting of labeled and unlabeled data samples.

\section{Acknowledgment}
The authors would like to thank Pascal Frossard and Alhussein Fawzi for the helpful discussions that contributed to this study.

\bibliographystyle{IEEEtran}
\bibliography{refs}

%

\end{document}